\documentclass[runningheads]{llncs}

\usepackage{eccv}

\usepackage{eccvabbrv}

\usepackage{graphicx}
\usepackage[export]{adjustbox}
\usepackage{booktabs}
\usepackage{multirow}

\usepackage[accsupp]{axessibility}  % Improves PDF readability for those with disabilities.

\usepackage[breaklinks,colorlinks,citecolor=eccvblue]{hyperref}
\usepackage{orcidlink}

\usepackage{multirow}
\usepackage[table]{xcolor}
\usepackage{silence}
\usepackage{microtype}
\usepackage{bibunits}
\makeatletter
\AtBeginDocument{%
  \providecommand*\@extra@binfo{}%
  \providecommand*\@extra@b@citeb{}%
  \def\bibcite#1#2{%
    \global\@namedef{b@#1\@extra@binfo}{%
      \hyper@@link[cite]{}{cite.#1\@extra@b@citeb}{#2}}%
  }%
}
\makeatother

\begin{document}
\begin{bibunit}[splncs04]

\title{Parallelized Autoregressive Decoding for Omni-Modal Dense Video Captioning} 

\titlerunning{PadCaptioner}

\author{Wenzheng Zeng, 
Siyi Jiao, 
Chen Gao, 
Hwee Tou Ng$^{\dagger}$, 
Mike Zheng Shou$^{\dagger}$} 

\authorrunning{W.~Zeng et al.}

\institute{National University of Singapore, Singapore \\
}

\maketitle
\let\thefootnote\relax\footnotetext[1]{$^{\dagger}$Corresponding authors.}

\begin{abstract}
Dense video captioning aims to generate temporally grounded descriptions of video events, benefiting both event-level video understanding and generation. In this domain, autoregressive video large language models have emerged as a prevalent paradigm due to their strong generative and cross-modal modeling capacity. However, generating dense captions under the token-by-token paradigm severely limits inference efficiency and hinders scalability as video length and event density increase. In this work, we propose a parallelized autoregressive framework that not only improves generation efficiency but also enhances temporally grounded captioning performance. Our key insight is to exploit the weak local dependencies across temporally distinct events to restructure the causal dependency graph, thereby enabling lossless parallel generation. Specifically, tokens with weak cross-event dependencies can be decoded in parallel, while tightly coupled tokens within each event retain sequential decoding to preserve local semantic coherence. 
To realize this insight, we introduce two key components for lossless parallel decoding: 
(1) a latent global planning mechanism that automatically learns the event-level structure and produces compact tokens encoding global inter-event causality while adaptively aggregating event-level audio-visual semantics, guiding subsequent dependency restructuring and parallel decoding; and 
(2) an event-factorized parallel decoding mechanism that effectively balances local focus with global inter-event awareness.
Experiments on various benchmarks demonstrate the clear advantage of our approach in both efficiency and performance in omni-modal event grounding and captioning. 
Project website: \url{https://github.com/showlab/PadCaptioner}.

\end{abstract}

\section{Introduction}
\label{sec:intro}

\begin{figure*}[t]
  \centering
  \includegraphics[width=\linewidth]{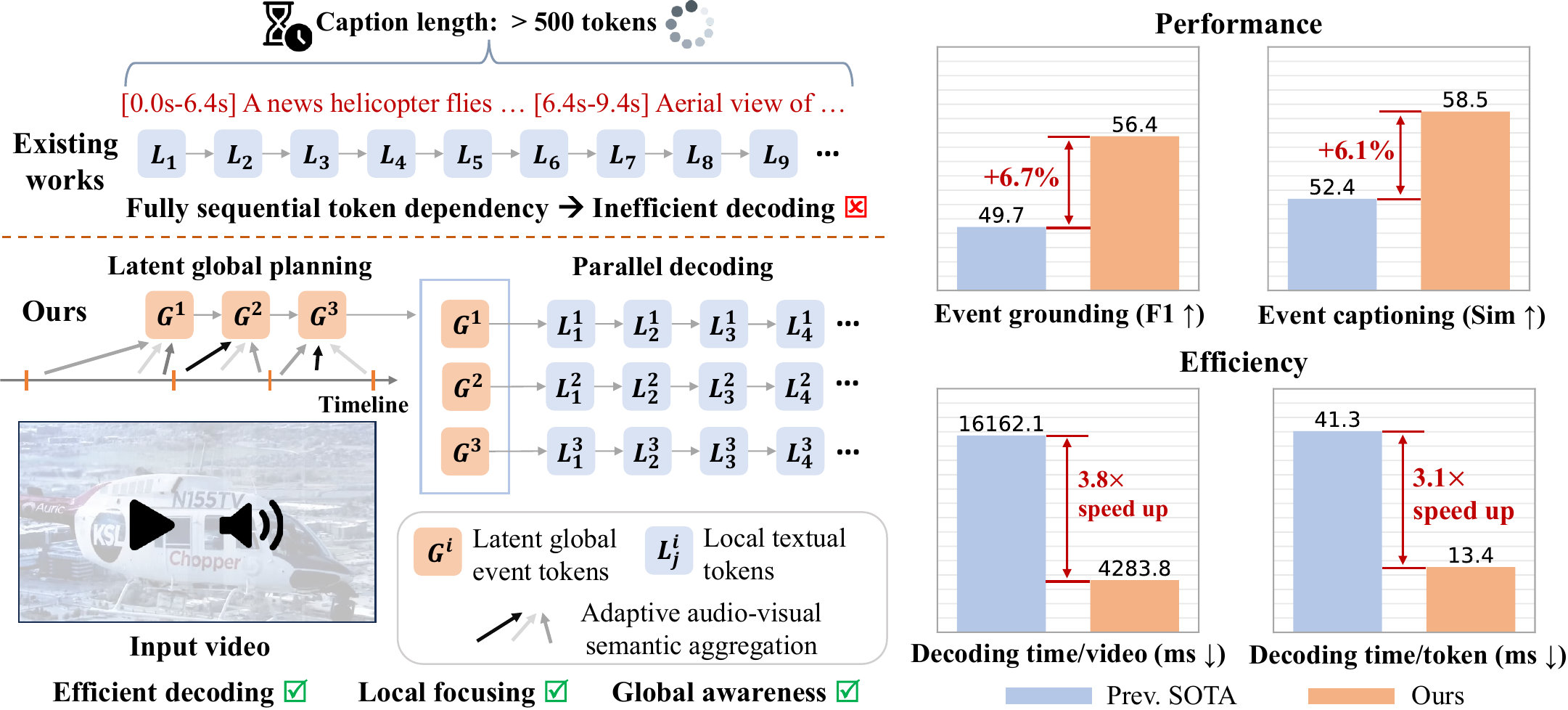} 
    \caption{\textbf{Left:} Comparison of token dependency modeling and decoding strategies between existing Video-LLM-based captioning models and ours. We achieve lossless parallel autoregressive decoding under a model-inferred restructured dependency graph. \textbf{Right:} Our method surpasses the previous SOTA~\cite{chronusomni_arxiv25} in both grounded captioning accuracy and decoding efficiency on the LongVALE benchmark~\cite{longvale}.}
\label{fig:fig1}
\end{figure*}

Dense Video Captioning (DVC) addresses the problem of describing multiple temporally localized events in untrimmed long videos~\cite{activitynet_caption_iccv17,youcook2_aaai18,longvale}. Unlike conventional video captioning, which typically produces a single global description, DVC requires jointly localizing densely occurring, semantically meaningful events and generating a temporally grounded caption for each event. This makes DVC substantially more challenging, yet highly valuable for applications such as egocentric perception~\cite{ego4d_cvpr22, ego_exo_4d_cvpr24,videollm_online_cvpr24}, video narration~\cite{livecc_cvpr25, streamingvlm_arxiv25, avocado_video_captioning_iclr26}, embodied agents~\cite{palm_e_embodied_vlm_agent_dvc_app_icml24,auroracap_iclr25}, video search and indexing~\cite{vid2seq_cvpr23}, and multi-event video generation~\cite{moviebench_cvpr25, moiveagent_arxiv25}.

To tackle DVC, autoregressive video large language models (Video-LLMs) have emerged as a dominant paradigm due to their strong generative capacity and cross-modal reasoning ability. By modeling video and language in a unified sequential framework, they provide a scalable solution for end-to-end event localization and caption generation. However, the strictly token-by-token decoding process incurs substantial latency. This issue is particularly severe in DVC, where each densely localized event requires generating a detailed multi-token description within a long-form video. As video length and event density increase, the total number of decoding steps grows rapidly, rendering sequential autoregressive decoding increasingly inefficient for long-form scenarios~\cite{growing_a_twig_iccv25}. Although recent diffusion-based LLMs (dLLMs) seek to accelerate generation through parallel decoding, they inherently face a dilemma in simultaneously matching both the efficiency and effectiveness of strong autoregressive models~\cite{d3llm_arxiv26}. This issue is even more pronounced in video understanding scenarios, which remains relatively underexplored by advanced diffusion-based models~\cite{dream_vl_video_dllm_arxiv25,vidlada_video_dllm_arxiv26}.

In this work, we propose PadCaptioner, a parallelized autoregressive framework that not only improves captioning efficiency but also enhances omni-modal temporal event grounding and caption quality. We argue that the sequential dependency chain modeled in autoregressive decoding frameworks is not strictly necessary for every token, and tokens with weak local dependencies can be predicted in parallel without sacrificing performance.
Based on this observation, our key insight is to exploit the weak local dependencies across temporally distinct events to restructure the causal dependency graph, thereby enabling lossless parallel generation. 
Specifically, tokens with weak cross-event dependencies can be decoded in parallel, while tightly coupled tokens within each event retain sequential decoding to preserve local semantic coherence.
However, naively applying such a strategy raises two fundamental challenges. First, identifying an appropriate event-level structure for dependency reformulation that enables valid parallel generation—particularly when learned automatically from raw video inputs—remains a challenge. Second, fully isolated event-wise parallel generation may overlook useful non-local dependencies among temporally distant events.

To address these challenges, we introduce a latent global planning mechanism that explicitly models non-local event-level structure before parallel decoding.
As shown in ~\cref{fig:fig1}, conditioned on the audio-video input, the model first autoregressively generates a sequence of compact global event tokens $\{G^1, G^2, \dots, G^K\}$. Each event token aims to ground and represent one temporally coherent event in the video. By generating these event tokens sequentially, the model preserves inter-event causal dependencies, forming a globally coordinated temporal structure. Meanwhile, each event token adaptively aggregates salient audio-visual information from its captured event, establishing enriched representations that facilitate subsequent event-wise decoding.

Guided by these globally coordinated event tokens, the model then transitions to an event-wise decoding phase. 
Rather than generating all caption tokens sequentially, we reformulate the token-level dependency graph according to the planned event-level structure, decomposing generation into multiple event-conditioned subchains. Each subchain, anchored by its corresponding global event token $G^i$, can be decoded in parallel across events. Meanwhile, within each subchain, sequential autoregressive generation is preserved to maintain strong intra-event dependencies and ensure local semantic coherence. At the same time, we adjust the inter-token visibility so that local tokens belonging to different subchains are prevented from attending to one another, eliminating unnecessary cross-event interference. Besides, each subchain retains full access to the shared global event tokens, allowing local event-wise generation to remain aware of global inter-event dependencies.

We conduct extensive experiments to evaluate the effectiveness of our framework. As shown in~\cref{fig:fig1}, our approach achieves SOTA performance on both event grounding and captioning, while delivering a $3.8\times$ actual wall-time decoding speedup. Besides, it also demonstrates strong generalization to other temporally grounded omni-modal video understanding tasks as verified by the experiments.

\section{Related Work}

% \subsubsection{Dense Video Captioning}
\noindent \textbf{Dense video captioning.}
Early dense video captioning (DVC) methods typically follow a proposal-and-caption paradigm, where temporal event segments are first localized and then described using sequence decoders~\cite{activitynet_caption_iccv17, jointly_dvc_cvpr18, masked_transformer_dvc_cvpr18, streamlined_dvc_two_stage_cvpr19}. Subsequent query-based approaches replace explicit proposal generation with learnable event queries, enabling joint event localization and caption prediction within a unified detection-style framework~\cite{pdvc_iccv21, dvc_query_based_cvpr24, dvc_query_based_acl25, e2dvc_cvpr25}.
More recently, DVC has increasingly adopted video large language models as the dominant paradigm~\cite{vid2seq_cvpr23, vtimellm,timechat_cvpr24, etbench_nips24, hicm2_aaai25, trace_iclr25, longvale,arc_chapter_arxiv25}. Benefiting from powerful pretrained backbones with strong language priors and cross-modal reasoning capabilities, these models achieve superior performance and exhibit broader task generalization compared to traditional task-specific DVC frameworks. However, their strictly sequential decoding process results in substantial inference latency, particularly due to the dense and multi-sentence generation nature of dense video captioning. In this work, we aim to improve not only the decoding efficiency but also the event grounding and captioning accuracy of prevalent autoregressive frameworks.

% \subsubsection{Parallel Prediction in Sequential Generation}
\smallskip
\noindent \textbf{Parallel prediction in sequential generation.}
Parallelization has been explored in various fields to improve efficiency of sequential generation. In language modeling, diffusion-based LLMs generate tokens through iterative refinement rather than strict left-to-right decoding~\cite{llada_arxiv25,dllm_survey_arxiv25}. Although these methods introduce position-wise parallelism, they typically require multiple refinement steps and cannot fully exploit KV cache acceleration, resulting in limited practical speedup over autoregressive (AR) decoding~\cite{block_diffusion_iclr25, d2f_dllm_arxiv25}. Moreover, they struggle to simultaneously match the efficiency and generation quality of strong AR-based models~\cite{d3llm_arxiv26}, especially in multimodal and video understanding scenarios~\cite{llada_v,dream_vl_video_dllm_arxiv25,vidlada_video_dllm_arxiv26}. 
In visual autoregressive generation, various methods~\cite{par_cvpr25, randar_cvpr25, nar_iccv25, ARPG_iclr26, lpd_iclr26} achieve parallel decoding by reorganizing dependencies over image tokens. They operate under a fixed spatial token layout with predetermined output cardinality, a structural assumption that does not hold in dense video captioning. Unlike visual generation, DVC requires producing variable-length textual outputs organized around dynamically evolving events, thereby introducing structural challenges that cannot be addressed by existing methods.

Within the DVC domain, conventional query-based methods such as PDVC~\cite{pdvc_iccv21} support parallelized event-level caption generation. However, their architectures substantially differ from the autoregressive transformer paradigm adopted by modern pretrained multimodal LLMs (MLLMs). As a result, they cannot inherit the rich world knowledge, response generation ability, and vision-language priors encoded in well-trained MLLMs, leading to suboptimal performance and limited generalization. 
In contrast, our work enables MLLM-native parallel caption generation through principled architectural repurposing, thereby naturally leveraging strong vision-language priors while addressing the inefficiency of modern autoregressive decoding.

% \subsubsection{Latent Reasoning}
\smallskip
\noindent \textbf{Latent reasoning.}
Latent reasoning augments sequence generation with intermediate latent tokens that enable structured reasoning before producing final outputs. Representative approaches generate intermediate visual representations in the latent space to facilitate grounded image~\cite{lira_seg_iccv25,padt_iclr26} and video perception~\cite{d2vlm_iccv25,meco_iclr26}, as well as long-horizon reasoning tasks~\cite{interleaved_latent_reasoning_arxiv25,sketch_in_latents_arxiv25,zebra_cot_iclr26,mirage_latent_reasoning_cvpr26}.
Our work employs latent planning to reason about event-level structure, with the goal of exploiting weak inter-event dependencies to restructure the conditional dependency graph, thereby enabling structured and lossless parallel decoding. During the latent planning stage, we also explore effective mechanisms to adaptively aggregate audio-visual semantics into the latent tokens to facilitate expressive decoding.

% \subsubsection{Temporally Grounded Multimodal LLMs}
\smallskip
\noindent \textbf{Temporally grounded multimodal LLMs.}
Temporally grounded multimodal LLMs aim to enhance time awareness and event-grounded perception in video-language understanding~\cite{timechat_cvpr24,vtimellm,etbench_nips24,vtgllm_aaai25,d2vlm_iccv25,timer1_arxiv25,timelens_cvpr26}. Most existing approaches are built upon visual-only architectures~\cite{timechat_cvpr24,momentor_icml2024,vtimellm,etbench_nips24,vtgllm_aaai25,trace_iclr25,d2vlm_iccv25,qwen25vl,qwen3vl_arxiv25,video_opd_temporal_grounding_arxiv26,videochat_flash, timer1_arxiv25, videochat_r1_arxiv25, videotg_r1_arxiv25, timesuite_iclr25}. Recently, several audio-visual (omni) LLMs have been proposed to incorporate acoustic cues for audio-visual perception~\cite{pandagpt,video_llama_emnlp23_demo,avicuna_aaai25, qwen25omni, qwen3_omni_arxiv25}. However, their temporal grounding capability remains unsatisfactory~\cite{longvale,trisense_nips25,chronusomni_arxiv25}.
In this work, we develop an omni-modal LLM that natively supports both visual and audio perception. While our primary focus is on improving both accuracy and decoding efficiency for dense video captioning through structured parallel autoregressive decoding, our framework also demonstrates strong generalization to other temporally grounded video understanding tasks, as validated by extensive experiments.

\section{Method}\label{sec:method}

\subsection{Preliminary}
We first outline the standard Omni-LLM architecture, which consists of modality-specific encoders and an LLM decoder. 
Given a video $\mathcal{V}$ with synchronized audio $\mathcal{A}$, the visual and audio streams are encoded into token sequences 
$Z_v \in \mathbb{R}^{N_v \times D}$ and 
$Z_a \in \mathbb{R}^{N_a \times D}$, 
where $N_v$ and $N_a$ correspond to the lengths of the visual and audio token sequences, respectively. $D$ is the hidden dimension of the LLM. 
To preserve temporal alignment, the video is partitioned into $T$ segments according to the visual frame sampling interval, where each segment corresponds to one sampled frame and its temporally aligned audio span.  Tokens within the same segment are grouped as multimodal chunks $C_t = [Z_v^{t}, Z_a^{t}]\in \mathbb{R}^{(N_v^t + N_a^t) \times D}$, yielding the sequence $F_M = \{C_t\}_{t=1}^T$. This multimodal sequence is then concatenated with the textual instruction embeddings $F_Q \in \mathbb{R}^{N_q \times D}$ to form the prefix $P$ for the Omni-LLM decoder. The model then generates output tokens autoregressively:
\begin{equation}
    L_i \sim \mathbb{P}(L_i \mid P, L_{<i}),
\label{eq:sequence_dependency}
\end{equation}
where each generated token $L_i$ is sampled from the distribution $\mathbb{P}$ generated by the LLM decoder, under the condition of prefix input $P$ and all previously generated tokens $L_{<i}$.

\begin{figure*}[t]
  \centering
  \includegraphics[width=\linewidth]{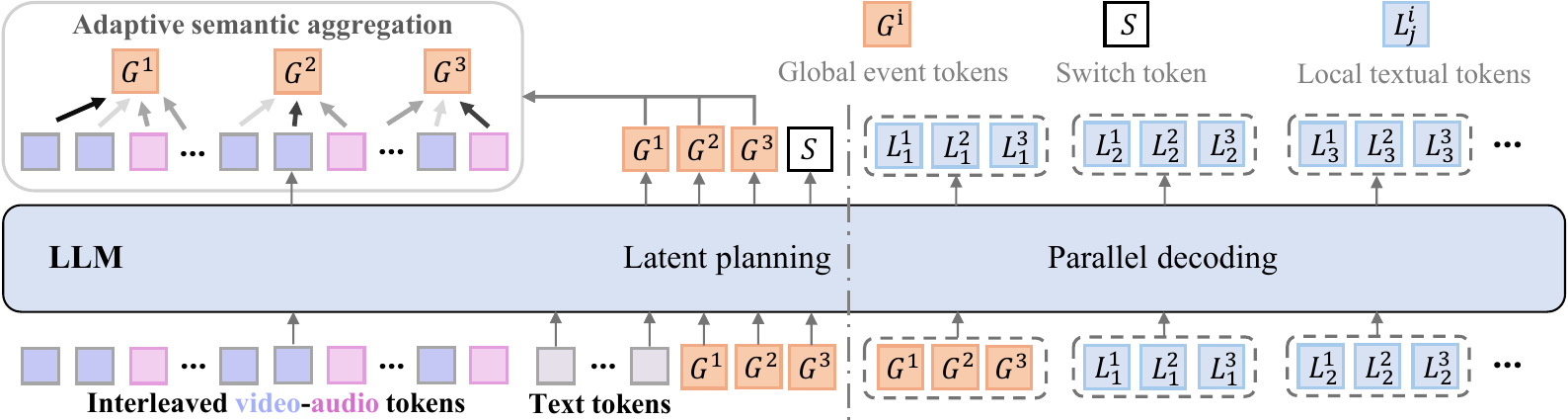}
    \caption{
    The overall pipeline of the proposed parallelized autoregressive decoding framework, where we set the model-inferred event count $K=3$ for illustration.}
\label{fig:pipeline}
\end{figure*}

\subsection{Method Overview}
In this work, we argue that the strictly sequential dependency chain (Eq.~\ref{eq:sequence_dependency}) adopted in existing frameworks is not uniformly necessary for every token, and that tokens with weak dependencies can be generated in parallel without sacrificing performance. Specifically, video content is inherently organized into temporally distinct events, providing a natural structural basis for dependency decomposition. Fine-grained semantic interactions are typically concentrated within event boundaries, while cross-event relations are largely governed by higher-level global semantics. Consequently, enforcing detailed token-level causal dependencies across events introduces redundant serialization, where such cross-event interactions can instead be conveyed through compact global representations that preserve inter-event coherence while relaxing unnecessary coupling.

Motivated by this observation, we propose a parallelized autoregressive framework that restructures the dependency graph to enable efficient parallel generation while maintaining or even strengthening the temporally grounded caption quality. As illustrated in~\cref{fig:pipeline}, the framework comprises two core components: (1) a latent global planning mechanism that discovers a valid event-level structure to guide dependency reformulation and aggregate event-centric audio-visual semantics for subsequent parallel decoding, and (2) a parallelized decoding strategy that balances local semantic coherence with global structural awareness, enabling lossless grounded caption generation.

% To enable structured parallel decoding, we introduce a latent global planning stage that models the event-level temporal structure of the video before token-level generation. From a dependency graph perspective, this stage constructs a compact set of global event anchors that facilitate the transformation of the fully causal token-level graph into event-specific subchains.
\subsection{Latent Global Planning} \label{sec: global_planning}

As shown in ~\cref{fig:pipeline}, conditioned on the multimodal prefix $P$, the model autoregressively generates a sequence of global event tokens $\{G^1, G^2, \ldots, G^K, S\}$, where each token $G^i$ is intended to encode a temporally coherent event as a condensed semantic representation in the latent space, and another special token $S$ will be generated at the end of the planning stage. Notably, the number of events $K$ is not predefined but adaptively inferred, enabling the model to capture the intrinsic temporal structure of each video rather than imposing a rigid and potentially unsuitable partition. This learned event-level abstraction decomposes the video into semantically coherent units prior to local textual token generation, establishing structured anchors that guide subsequent dependency graph reformulation and parallel decoding.

Technically, the latent global token $G^i$ is implemented by introducing a special token \texttt{<G>} into the vocabulary, which the model is trained to autoregressively predict during the planning stage. 
To ensure that the planned event-level abstraction provides structurally valid and semantically sufficient anchors for subsequent parallel decoding, we introduce the following mechanisms.

\subsubsection{Explicit Event Grounding Constraint}
For the planned event-level structure to effectively support subsequent dependency graph reformulation and parallel decoding, the generated global event tokens should be temporally grounded. In the dense video captioning task, the annotated ground-truth event intervals naturally define a temporally grounded semantic partition of the video. We leverage this structural prior during training to guide the learning of global event tokens, encouraging them to align with meaningful event-level units.

Specifically, inspired by feature matching-based grounding methods~\cite{etbench_nips24, d2vlm_iccv25}, for each generated global event token $G^i$, we impose a similarity-based grounding constraint that aligns its representation with multimodal (audio-visual) prefix features within its corresponding ground-truth temporal segment, while suppressing similarity to features outside this interval. Formally, we optimize:

\begin{equation}
    \mathcal{L}_{\text {gnd}}^{G^i}=\frac{1}{T} \sum_{t=1}^T \operatorname{BCE}\left(y^i_t, \operatorname{sim}^i_t\right)
\end{equation}
where $\operatorname{BCE}$ denotes the binary cross-entropy loss. 
$y^i_t \in \{0,1\}$ is the binary indicator specifying whether time segment $t$ 
lies within the $i$-th ground-truth event interval, 
and $\mathrm{sim}^i_t$ is computed as the normalized dot-product similarity 
between $G^i$ and the corresponding multimodal tokens within the corresponding time segment.
This explicit grounding constraint enforces each global event token to align with a temporally valid event segment, thereby establishing reliable anchors for subsequent dependency graph reformulation. Furthermore, it equips the generated $G_i$ tokens with intrinsic grounding capability, such that event localization can be effectively performed at inference time through similarity matching with the multimodal prefix tokens.

\subsubsection{Adaptive Semantic Aggregation}

Although the grounding constraint ensures the temporal validity of $G^i$, we expect the global event tokens to play a more substantial role. Beyond determining the event decomposition for parallel generation, each $G^i$ should also encode sufficiently expressive and event-specific semantics. Such semantic richness enables the event token to function as a meaningful anchor in the subsequent parallel decoding stage, allowing each parallelized branch to focus on its designated event rather than being diluted by globally diffused context.
The importance of semantically rich representations has also been observed in general image/video grounding and perception literature~\cite{lira_seg_iccv25,d2vlm_iccv25}.
In our framework, the semantic expressiveness of the global event tokens plays a critical role in stabilizing parallel branch-wise generation and reinforcing event-level focus.

Technically, we aim to aggregate the event-relevant semantics encoded in the audio-visual prefix tokens into the corresponding global event tokens. For each generated event token $G^i$, we consider the prefix tokens that fall within its temporally grounded time segment. Guided by this principle, we explore and discuss three strategies for effectively aggregating semantics from dense audio-visual prefix tokens.

\noindent \textbf{(1) Uniform aggregation.} 
As a straightforward aggregation strategy, we apply mean pooling over the selected salient tokens and add the pooled feature to the embedding of $G^i$. This approach aggregates information uniformly and provides a simple yet stable way to integrate token-level semantics.

\noindent \textbf{(2) Adaptive aggregation through learned scoring.}
We further consider an adaptive aggregation strategy based on learned importance scores. The motivation is that tokens across space, time, and modality are not equally informative and expressive for representing the underlying event. To capture this heterogeneity, we build a lightweight scoring head, instantiated as a two-layer MLP on top of the latent token embeddings, which predicts a scalar importance score for each salient token. These scores are then used to perform weighted pooling, producing an aggregated feature that is added to $G^i$.

\noindent \textbf{(3) Attention-guided aggregation.}
As an alternative adaptive strategy, we derive token importance directly from the attention map between the text query tokens and the multimodal tokens. This approach shares a similar motivation with the learned score head but avoids introducing additional parameters. The key intuition is that the attention mechanism naturally focuses on task-relevant and informative tokens; therefore, the attention weights can serve as an effective proxy for token importance when performing weighted aggregation. In practice, we use the attention weights between the last text query token and the multimodal prefix tokens to compute importance scores, as it typically summarizes the full query context in autoregressive modeling.
% We empirically find that this simple and parameter-free strategy is more effective than the learned scoring head.

\subsection{Dependency-Restructured Parallel Decoding}\label{sec: parallel_decoding}

Guided by the globally coordinated event tokens produced during the planning stage, we reformulate the original fully serialized token dependency into a structured, event-factorized dependency graph. Specifically, instead of modeling caption generation as a single causal chain over all tokens, we decompose it into multiple weakly coupled, event-conditioned subchains so that different subchains can be decoded in parallel. Importantly, our objective is to balance two seemingly competing requirements:
(i) preserving strong event-focused autoregression within each branch while preventing unnecessary cross-event local interactions that may distract event-specific decoding; and
(ii) maintaining global awareness of inter-event structure across branches.

To achieve this, we adopt a factorized dependency structure where tokens from different events are rendered conditionally independent given the shared global context and local histories, enabling synchronous parallel decoding across branches. Specifically, at each decoding step $j$, local tokens across all $K$ event branches are decoded simultaneously under the conditional distribution:

\begin{equation}
\left\{ L_{j}^{i} \right\} _{i=1}^{K}\sim \mathbb{P} (\left\{ L_{j}^{i} \right\} _{i=1}^{K}\mid P,G^{1:K},\left\{ G^i,L_{<j}^{i} \right\} _{i=1}^{K}),
\end{equation}
where $P$ denotes the shared multimodal prefix, and $\{G^1,\dots,G^K\}$ are the global event tokens generated in Sec.~\ref{sec: global_planning}, which encode global inter-event structure. At the beginning of subchain decoding, the corresponding event token $G^i$ will serve as a conditional anchor that guides event-focused generation, as illustrated in ~\cref{fig:pipeline}.
Despite adopting an event-wise parallel decoding paradigm, our approach preserves the autoregressive property within each subchain to maintain strong local semantic coherence.

\begin{figure*}[t]
  \centering
  \includegraphics[width=0.98\linewidth]{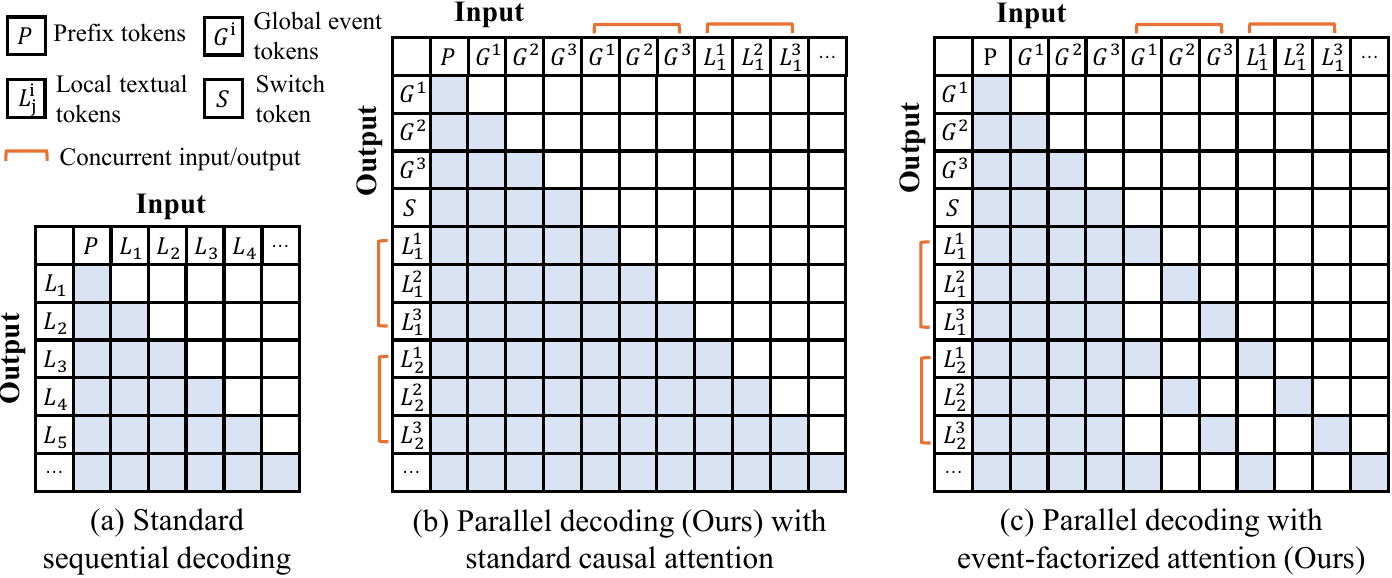}
    \caption{
    Comparison of the different decoding strategies and attention mechanisms.}
\label{fig:attention}
\end{figure*}

Based on the reformulated dependency graph, we derive a modified causal attention scheme that enables structured parallel decoding. As shown in ~\cref{fig:attention}, during local generation, each token $L^i_j$ is permitted to attend to: (1) the shared multimodal prefix $P$ and all global event tokens $\{G^1,\dots,G^K\}$, ensuring global inter-event awareness, and (2) the event anchor token $G^i$, as well as previously generated local tokens $\{L^i_{<j}\}$ within the same branch, preserving strong intra-event autoregression. Meanwhile, tokens from different event subchains are explicitly masked from attending to one another, thereby eliminating unnecessary cross-event dependencies. Empirical results in~\cref{sec:abla} demonstrate that this factorized attention effectively balances local and global dependencies under parallel decoding. In contrast, naive causal attention introduces spurious cross-event interactions and suffers significant performance degradation.

Notably, event-level responses can inherently vary in length across different sub-chains. To accommodate this variability while preserving synchronized parallel decoding, we introduce two complementary design strategies.
First, we modify the positional encoding scheme such that tokens at the same decoding step across different subchains share identical position indices. This design reinforces the structural independence among subchains and enables efficient parallel inference: once a subchain generates an end-of-sequence (EOS) token, it can terminate independently without affecting other active branches, thereby avoiding unnecessary generation of redundant tokens. Second, during training, we adopt a padding-based alignment strategy. Specifically, token subchains shorter than the maximum length $N$ (defined as the largest token length among the $K$ branches) are padded with additional EOS tokens. We involve the training loss on predicting these padded EOS tokens, which we empirically find beneficial for performance, as this likely encourages each subchain to terminate promptly rather than generating redundant tokens. Notably, this padding scheme is straightforward to implement and remains fully compatible with standard autoregressive transformer training. Thanks to the aforementioned positional encoding design and the subchain termination mechanism, the padding introduced during training does not affect inference (i.e., no redundant tokens will be generated during inference).

\section{Experiments}

\subsection{Experimental Setup}

% \subsubsection{Implementation Details}
\noindent \textbf{Implementation details.}
We adopt the recent state-of-the-art omni-LLM, Video-SALMONN 2+ (3B)~\cite{video_salmonn2}, as our base model. The model employs Qwen2.5-VL~\cite{qwen25vl} as both the visual encoder and the LLM decoder, and utilizes Whisper-Large-v3~\cite{whisper_v2_speech_encoder} as the audio encoder, with a window-level Q-Former~\cite{audio_qformer_icassp24} serving as the audio aligner. 
Following~\cite{chronusomni_arxiv25}, we fine-tune the model on a subset of its training data (18K videos from ChronusAV~\cite{chronusomni_arxiv25}) using LoRA~\cite{lora} for one epoch. Beyond the constraints introduced in~\cref{sec:method}, we apply token-level classification loss for standard supervised fine-tuning under the restructured sequence (\cref{fig:pipeline}).
We adopt a frame sampling rate of 0.5 FPS for both training and inference, with a maximum of 256 frames. If a video exceeds this limit, frames are uniformly sampled to obtain 256 frames.
More details can be found in the appendix.

\smallskip

\noindent \textbf{Evaluation tasks and benchmarks.}
We evaluate our method on multiple dense video captioning (DVC) benchmarks, as well as other time-sensitive video understanding tasks, to assess its generalization capability.
Specifically, we conduct DVC experiments on omni-modal benchmarks LongVALE~\cite{longvale} and ChronusAV~\cite{chronusomni_arxiv25}, as well as the visually-centric benchmark YouCook2~\cite{youcook2_aaai18}, following their standard evaluation protocols. Since ChronusAV does not report results on the DVC task, we additionally conduct comparative experiments using the official checkpoints of existing methods. 
Following~\cite{etbench_nips24}, we also report F1 score to specifically evaluate event grounding accuracy, and sentence similarity (Sim) to measure the semantic similarity between model outputs and ground-truth captions.
Beyond DVC, we further evaluate our approach on additional time-sensitive tasks, including omni-modal temporal grounding and segment-level captioning tasks from the LongVALE (2 tasks) and ChronusAV (6 tasks) benchmarks.

\begin{table}[t]
\centering
\footnotesize
\caption{Dense video captioning results on LongVALE and ChronusAV. ``PadCaptioner$^-$'' means our model trained on a smaller training budget (12K videos). C: CIDEr. S: SODA\_c. M: METEOR.}
\label{tab:omni_dvc}
\begin{tabular}{l|ccccc|ccccc}
\toprule
\multirow{2}{*}{Method} & \multicolumn{5}{c|}{LongVALE} & \multicolumn{5}{c}{ChronusAV} \\
                        & F1 & Sim & S& C & M & F1 & Sim & S & C & M \\ 
\midrule
TimeChat (7B)~\cite{timechat_cvpr24} & - & - & 1.6 & 0.1 & 1.4 & - & - & - & - & - \\
LongVALE-LLM (7B)~\cite{longvale} & 31.2 & 37.2 & 2.8 & 7.9 & 4.7 & 18.4 & 25.4 & 3.1 & 4.6 & 8.8 \\
video-SALMONN $2+$ (3B)~\cite{video_salmonn2} & 35.5 & 26.6 & 1.9 & 2.4 & 2.4 & 21.5 & 17.6 & 2.8 & 1.5 & 4.3 \\
Qwen3-Omni (30B-A3B)~\cite{qwen3_omni_arxiv25} & 34.2 & 45.8 & 4.2 & 5.7 & 5.5 & 51.8 & 37.0 & 5.4 & 2.5 & 6.5 \\
ChronusOmni (7B)~\cite{chronusomni_arxiv25} & 49.7 & 52.4 & 3.7 & 5.6 & 5.2 & 60.1 & 36.8 & 8.6 & 2.9 & 7.6 \\
\rowcolor[HTML]{EDEDFF}
PadCaptioner$^{-}$ (3B) & 53.0 & 55.5 & 5.6 & 12.3 & 7.8 & 61.8 & 38.8 & 10.5 & 7.8 & 10.6 \\
\rowcolor[HTML]{EDEDFF}
PadCaptioner (3B) & \textbf{56.4} & \textbf{58.5} & \textbf{6.4} & \textbf{13.7} & \textbf{8.6} & \textbf{63.2} & \textbf{40.0} & \textbf{12.4} & \textbf{9.6} & \textbf{12.4} \\
\bottomrule
\end{tabular}
\end{table}

\subsection{Main Results on Dense Video Captioning}

\noindent \textbf{LongVALE.}
As shown in the left part of~\cref{tab:omni_dvc}, PadCaptioner outperforms recent SOTA approaches by large margins across all metrics (e.g., at least 6.7\% in F1 for event grounding and 6.1\% in Sim for event captioning). Moreover, our model is relatively lightweight compared with most existing methods that employ 7B-scale models. It also achieves substantially faster decoding, as analyzed later. These results demonstrate the clear advantages of our approach in both temporally grounded captioning accuracy and efficiency.

\smallskip
\noindent \textbf{ChronusAV.}
As shown in the right part of~\cref{tab:omni_dvc}, PadCaptioner maintains strong performance on ChronusAV, further demonstrating the effectiveness and generalization ability of our framework for omni-modal dense video captioning.

\begin{table}[t]
\centering
\scriptsize
\caption{Efficiency analysis on the dense video captioning task on LongVALE and ChronusAV. 
T/video and T/token denote the average decoding latency per video and per token, respectively, measured by actual wall-clock time (in milliseconds).}
\label{tab:efficiency}
\begin{tabular}{l|cccc|cccc}
\toprule
\multirow{2}{*}{Method} 
& \multicolumn{4}{c|}{LongVALE} 
& \multicolumn{4}{c}{ChronusAV} \\
\cmidrule(lr){2-5} \cmidrule(lr){6-9}
& F1$\uparrow$  & Sim$\uparrow$  & T/video$\downarrow$ & T/token$\downarrow$ 
& F1$\uparrow$  & Sim$\uparrow$  & T/video$\downarrow$ & T/token$\downarrow$\\
\midrule
video-SALMONN-2+ (3B)~\cite{video_salmonn2} & 35.5 & 26.6 & \textbf{2706.8} & 22.5 & 21.5 & 17.6 & \textbf{3083.4} & 22.8 \\
ChronusOmni (7B)~\cite{chronusomni_arxiv25}& 49.7 & 52.4 & 16162.1 & 41.3  & 60.1 & 36.8 & 28093.0 & 41.0 \\
\rowcolor[HTML]{EDEDFF}
PadCaptioner (3B) & \textbf{56.4} & \textbf{58.5} & 4283.8 & \textbf{13.4} & \textbf{63.2} & \textbf{40.0} & 7526.7 & \textbf{13.7} \\
\bottomrule
\end{tabular}
\end{table}

\smallskip
\noindent \textbf{Efficiency analysis.}
We compare our method with ChronusOmni (7B)~\cite{chronusomni_arxiv25}, the strongest recent omni-modal LLM, as well as a 3B base model~\cite{video_salmonn2}. All performance is measured on a single NVIDIA A6000 GPU. As shown in~\cref{tab:efficiency}, our method achieves clear advantages over the previous SOTA in both captioning performance and decoding efficiency (i.e., around 3.7× speedup in total decoding time and 3× speedup in decoding speed per token on both datasets). 
Although the total decoding time of our model is longer than that of the 3B counterpart, this is mainly because the latter tends to generate much shorter captions that lack expressive and informative content, which is also reflected in its significantly lower evaluation scores (e.g., F1, Sim). When normalized by the number of generated tokens, our method demonstrates a clear advantage in decoding efficiency.

\smallskip
\noindent \textbf{YouCook2.} 
We further evaluate our model on the visually-centric dense video captioning benchmark YouCook2 (\cref{tab:youcook2}). Our method achieves strong performance under zero-shot inference, further demonstrating its superiority.

\begin{table*}[t]
\centering
\scriptsize
\begin{minipage}[t]{0.36\textwidth}
\centering
\caption{Zero-shot evaluation on YouCook2. C: CIDEr. S: SODA\_c.}
\label{tab:youcook2}
\begin{tabular}{lccc}
\toprule
Method             & F1 & C & S \\ 
\midrule
TimeChat (7B)~\cite{timechat_cvpr24}        & 12.6  & 3.4     & 1.2      \\
VTimeLLM (7B)~\cite{vtimellm}        & 17.5  & 5.0     & 1.5      \\
TRACE (7B)~\cite{trace_iclr25} & 22.4  & 8.1     & 2.2      \\
D$^2$VLM (4B)~\cite{d2vlm_iccv25} & 26.4  & 10.6     & 3.2      \\
TimeExpert (6B)~\cite{timeexpert_iccv25} & 23.6  & 8.2     & 2.5      \\
ChronusOmni (7B)~\cite{chronusomni_arxiv25}     & 24.2  & 27.3     & 6.8      \\
\rowcolor[HTML]{EDEDFF} 
PadCaptioner (3B)            & \textbf{27.1}  & \textbf{31.7}     & \textbf{8.2}      \\
\bottomrule
\end{tabular}
\end{minipage}
\hfill
\begin{minipage}[t]{0.6\textwidth}
\centering
\caption{Omni-TVG and Omni-SC results on LongVALE. B: BLEU-4. R: ROUGE-L. C: CIDEr. M: METEOR.}
\label{tab:longvale_other_tasks}
\begin{tabular}{l|ccccc}
\toprule
\multirow{2}{*}{Method} & Omni-TVG & \multicolumn{4}{c}{Omni-SC} \\
                        & mIoU     & B     & R     & C     & M        \\ 
\midrule
VideoChat (7B)~\cite{videochat}            & 3.0        & 0.5     & 9.6     & 0.0     & 8.2    \\
VideoChatGPT (7B)~\cite{video_chatgpt_acl24}            & 5.0        & 0.4     & 14.0     & 0.9     & 5.9    \\
LongVALE-LLM (7B)~\cite{longvale}           & 11.0        & 5.6     & 22.4     & 20.3     & 10.9    \\
Trisense (7B)~\cite{trisense_nips25}            & 11.2        & 4.8     &   21.9   &  18.8    & 10.4    \\
ChronusOmni (7B)~\cite{chronusomni_arxiv25}          & 34.5        & 5.5     & 22.0     & 20.3     & 11.7    \\
\rowcolor[HTML]{EDEDFF} 
PadCaptioner (3B)                 & \textbf{45.7}        & \textbf{8.7}     & \textbf{26.0}     & \textbf{22.3}     & \textbf{12.4}    \\
\bottomrule
\end{tabular}
\end{minipage}
\end{table*}

\begin{table*}
  \centering
    \caption{Results on six temporally grounded video-audio understanding tasks in ChronusAV benchmark. B: BLEU-4. R: ROUGE-L. M: METEOR. C: CIDEr.}
  \resizebox{1\linewidth}{!}{ 
\begin{tabular}{l|cc|cccc|cc|cccc|cccc|cccc}
    \toprule
    \multirow{2}{*}{Model}  &
    \multicolumn{2}{c}{V2T} & \multicolumn{4}{c}{T2V} & \multicolumn{2}{c}{A2T} & \multicolumn{4}{c}{T2A} & \multicolumn{4}{c}{V2A}& \multicolumn{4}{c}{A2V}\\
    \cmidrule(lr){2-21} 
     &  R@0.5 & R@0.7 & B& R&M & C & R@0.5 & R@0.7 & B& R& M & C & B& R& M & C & B& R& M & C \\
    \midrule
    VideoLLaMA (7B)~\cite{video_llama_emnlp23_demo} &2.0&0.8&0.1&1.5&0.9&0.0 &1.7&0.8&0.1&1.1 & 0.6 &0.1 &0.1 &1.1&0.8 &0.1  &0.1 &1.2 &0.9 &0.0\\
    Ola (7B)~\cite{ola_arxiv25} &5.8&2.7&0.3&1.7&1.0&0.5 &5.5&2.6 &0.2&0.8&0.3&0.3 &0.2&1.2&0.5&0.5  &0.2 &1.6 &1.1& 0.4\\
    AVicuna (7B)~\cite{avicuna_aaai25}  &10.8 &4.7  & 0.0 &1.0 &0.3 &0.1 &8.2 & 3.5  & 0.0 & 0.5 &0.1 &0.0 &0.0 &0.5 &0.2 &0.1 &0.1 &1.5 &0.7 & 0.4\\
    LongVALE-LLM (7B)~\cite{longvale} &9.5&3.7 &0.4&2.0&1.0&1.4 &4.3 & 1.3 &0.2 &1.3 &0.5 &0.2 & 0.1 &1.3 &0.6 &0.2 &0.2 &1.9 &1.0&0.9\\
    Qwen2.5-Omni (7B)~\cite{qwen25omni} &7.1&3.0  &0.4 &1.9& 1.0 &0.8 &10.1 &3.7 &0.7 &1.0&0.6&0.3&0.5&0.9&0.6&0.3 &0.2&1.6&1.0&0.6\\ 
    ARC-HY-Video (7B)~\cite{arc_hunyuan_video_arxiv25}&36.1 &23.2 &0.2 &1.5&1.1&0.5 &36.9 &24.3 &0.2&1.2&0.7&0.4 & 0.1 & 0.9 & 0.7 &0.1 &0.1 &1.1 & 0.9 & 0.1\\  
    Qwen3-Omni (30B-A3B)~\cite{qwen3_omni_arxiv25}  &37.9&21.8 &0.4&2.1&1.6&0.9 &46.7 &33.1 &0.9&2.3&1.2&2.2 &0.4 &1.6 &1.3 &0.7 &0.2 &1.7&1.5&0.4\\
    ChronusOmni (7B)~\cite{chronusomni_arxiv25}  &63.2 &46.0  & 1.2 &3.4 & 2.1 & 5.1 &90.5 & 79.9  &6.8&6.9&4.5&34.3&3.6&4.9&3.3&13.6 & 1.0 & 3.2 & 2.1 & 3.0\\
    \rowcolor[HTML]{EDEDFF} 
    PadCaptioner (3B)  &\textbf{68.4} &\textbf{48.7}  & \textbf{1.4} &\textbf{3.9} & \textbf{2.7} & \textbf{5.6} &\textbf{91.3} & \textbf{82.5}  &\textbf{7.4}&\textbf{7.3}&\textbf{4.8}&\textbf{35.1}&\textbf{3.9}&\textbf{5.3}&\textbf{3.8}&\textbf{15.0} & \textbf{1.4} & \textbf{3.7} & \textbf{2.6} & \textbf{3.7}\\
    \bottomrule
  \end{tabular}}
  \label{tab:chronus}
\end{table*}

\subsection{Generalization to Other Temporally Grounded Tasks}

\noindent \textbf{LongVALE.} 
Here we evaluate the tasks in LongVALE beyond DVC, namely Omni-TVG for temporal audio-visual event grounding and Omni-SC for audio-visual segment captioning.
As shown in~\cref{tab:longvale_other_tasks}, our method performs consistently well on both tasks, indicating its generalization and broad applicability across diverse temporally grounded audio-video scenarios beyond DVC.

\smallskip
\noindent \textbf{ChronusAV.} 
Here we evaluate the six tasks defined in ChronusAV, which emphasize both time awareness and cross-modal reasoning~\cite{chronusomni_arxiv25}.
% We refer readers to the original paper for detailed task definitions.
From~\cref{tab:chronus}, our method also demonstrates competitive performance across all tasks, further validating the robustness and generalization capability of our framework for omni-modal temporally grounded video understanding.

\begin{table*}[t]
\scriptsize
\centering
\begin{minipage}[t]{0.56\textwidth}
\centering
\caption{Global effect of our designs.}
\label{tab:global_effect}
\begin{tabular}{l|cc|cc}
\toprule
Setting                          & F1$\uparrow$ & Sim$\uparrow$ & T/video$\downarrow$ & T/token$\downarrow$ \\
\midrule
Baseline                         & 32.7  & 17.6 & 7798.7  & 22.9  \\
+ planning                       & 61.5  & 38.4 & 12405.1  & 22.9  \\
++ parallel decoding             & \textbf{61.8}  & \textbf{38.8} & \textbf{7598.2}  & \textbf{13.8}  \\
\bottomrule
\end{tabular}
\end{minipage}
\hfill
\begin{minipage}[t]{0.42\textwidth}
\centering
\caption{Effect of latent planning.}
\label{tab:latent_planning}
\begin{tabular}{l|cc}
\toprule
Setting             & F1 & Sim \\
\midrule
Baseline            & 32.7  & 17.6 \\
+ Textual planning  & 37.4  & 19.4 \\
+ Latent planning   & \textbf{61.5}  & \textbf{38.4} \\
\bottomrule
\end{tabular}
\end{minipage}
\end{table*}

\begin{table*}[t]
% \footnotesize
\centering
\begin{minipage}[t]{0.48\textwidth}
% \footnotesize
\centering
\caption{Effect of attention mechanism.}
\label{tab:attention}
{\fontsize{9}{10}\selectfont
\begin{tabular}{l|cc}
\toprule
Setting           & F1 & Sim \\
\midrule
Causal attention        & 38.7  & 19.9 \\
Ours (self G-token only) & 59.8 & 37.6 \\
Ours (full G-token access) & \textbf{61.8}   & \textbf{38.8} \\
\bottomrule
\end{tabular}}
\end{minipage}
\hfill
\begin{minipage}[t]{0.48\textwidth}
% \footnotesize
\scriptsize
\centering
\caption{Aggregation strategy ablation.}
\label{tab:aggregation_strategy}
{\fontsize{8}{9}\selectfont
\begin{tabular}{l|cc}
\toprule
Setting             & F1 & Sim \\
\midrule
w/o aggregation     & 47.7   & 24.3 \\
Mean pooling        & 58.5   & 35.7 \\
Scoring head        & 61.4   & 38.2 \\
Attention guided    & \textbf{61.8}   & \textbf{38.8} \\
\bottomrule
\end{tabular}}
\end{minipage}
\end{table*}

\subsection{Ablation Studies} \label{sec:abla}
We conduct ablation studies on the ChronusAV dataset to assess the effect of each proposed component on both event grounding and captioning. All models in this section are trained on 12K videos for fair comparison.
Additional analyses are provided in the appendix.

\smallskip
\noindent \textbf{Effect of latent planning.}
(1) As shown in \cref{tab:global_effect}, latent planning substantially improves DVC performance in terms of both grounding and caption quality. Importantly, this design also provides valid event-level structure, making the subsequent dependency restructuring and parallel decoding possible.
(2) Interestingly, \cref{tab:latent_planning} shows that pure textual planning also brings a noticeable performance gain, where event localizations are first generated in textual form and then used to guide caption generation. This suggests that explicit planning is generally beneficial for long-context, dense event-level video understanding. Compared with textual planning, our latent planning achieves much stronger performance, which we attribute to its compact latent representation that captures event-level semantics more faithfully and provides more expressive anchors for event-focused caption generation.

\smallskip
\noindent \textbf{Effect of parallel decoding.}
As shown in \cref{tab:global_effect}, the proposed parallel decoding strategy significantly accelerates decoding (Row-2 vs. Row-3), achieving about $1.63\times$ overall speedup and $1.66\times$ per-token speedup.
Although Row-3 only slightly reduces the total wall-clock time over the baseline (Row-1), this is mainly because the baseline generates much shorter and less informative captions, as reflected by its much lower F1 and Sim scores. After normalizing by generated tokens (T/token), our method still achieves a $1.66\times$ decoding speedup.
Moreover, comparing Row-2 and Row-3 shows that parallel decoding maintains and even slightly improves DVC performance. We attribute this to the event-wise decoding design, which encourages event-focused caption generation and reduces interference from long-context dependencies and redundant cross-event information. 
Notably, such parallelized decoding cannot work effectively without the proposed factorized attention mechanism, as further analyzed in the following subsection.

\smallskip
\noindent \textbf{Effect of attention mechanism.}
We study the proposed event-factorized attention mechanism in \cref{tab:attention}. In Row-2 (``self G-token only''), besides causal local token visibility, each branch can only access its own global token $G^i$, while Row-3 uses our full design, where each branch can access all global event tokens.
From \cref{tab:attention}, we make two observations.
(1) Row-1 vs. Row-3: Standard causal attention causes a significant performance drop. This is because, under parallel decoding, each branch is conditioned on weakly related local tokens from other events, which introduces interference and weakens event-focused decoding.
(2) Row-2 vs. Row-3: Allowing full visibility of all global event tokens brings consistent improvements. This suggests that global event tokens effectively summarize inter-event relationships, allowing cross-event interactions to be conveyed through compact global representations, thereby preserving global coherence without unnecessary coupling between event-specific decoding branches.
Overall, event-factorized attention better balances local and global dependencies during parallel decoding, enabling faster inference while even improving captioning quality.

\begin{figure*}[t]
  \centering
  \includegraphics[width=\linewidth]{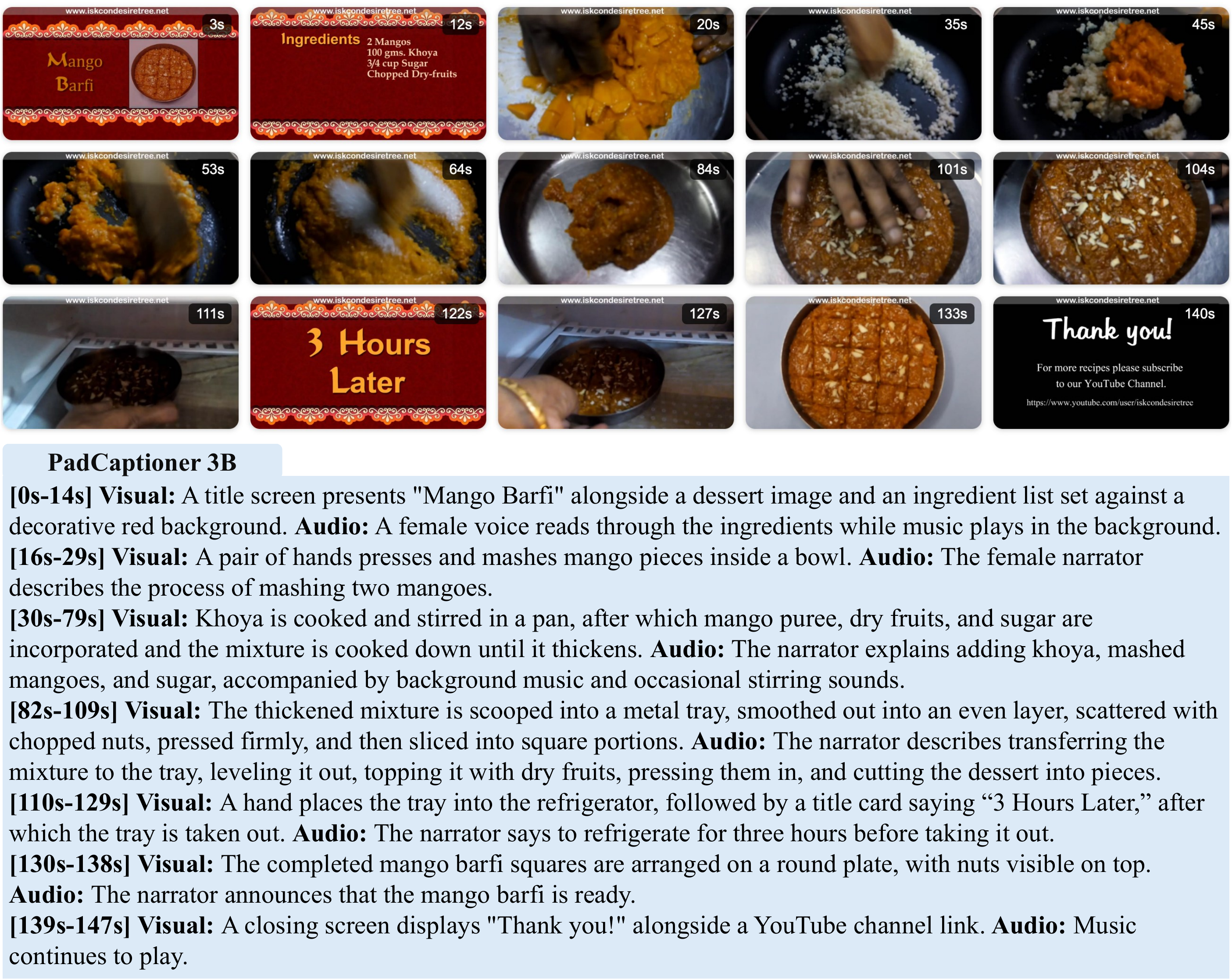}
    \caption{
    Qualitative example of a video caption generated by PadCaptioner.}
\label{fig:dvc_main}
\end{figure*}

\smallskip
\noindent \textbf{Effect of adaptive semantic aggregation.}
(1) As shown in~\cref{tab:aggregation_strategy}, removing audio-visual semantic aggregation into global event tokens (Row-1 vs. the remaining rows) leads to a substantial performance drop. This highlights the benefit of enriching global event tokens with event-focused semantics, enabling them to serve as expressive anchors for subsequent event-focused decoding.
(2) From Row-2 to Row-4: We investigate the effectiveness of different semantic aggregation strategies. (i) Adaptive strategies (Row-3 and Row-4) outperform rigid mean pooling (Row-2), suggesting that modeling relative token-level importance across space, time, and modality helps aggregate informative semantics from dense prefix tokens; 
(ii) Attention-guided aggregation (Row-4) outperforms the learned scoring head (Row-3), showing that internal attention weights can effectively estimate token importance without introducing additional parameters.

\subsection{Qualitative Analysis}

% We visualize a caption generated by PadCaptioner in~\cref{fig:dvc_main}. Our model can reasonably split and localize video events and generate informative descriptions that consider both visual and audio cues. More cases and analyses can be found in the appendix.
We show a caption generated by PadCaptioner in~\cref{fig:dvc_main}. Our model can reasonably segment and localize video events, while generating informative descriptions that leverage both visual and audio cues. More examples and analyses are provided in the appendix.

\section{Conclusion and Limitations}

In this work, we observe that prior Video-LLM-based dense video captioning models suffer from inefficient decoding due to their strictly sequential generation paradigm, which introduces unnecessary token-level dependencies across temporally distinct events. 
Based on this observation, we propose PadCaptioner, a parallelized autoregressive framework that restructures the dependency graph through latent event-level planning, accompanied by a parallel decoding mechanism that effectively balances local and global dependencies.
Extensive experiments demonstrate that PadCaptioner significantly improves inference efficiency and temporally grounded captioning accuracy, while also showing strong generalization to other temporally grounded audio-visual understanding tasks.

Our work still has several limitations. (1) Our dependency-graph restructuring considers only event-level structures. Future work could explore finer-grained cues (e.g., sub-events or spatiotemporal instances) and learn parallelizable units with less reliance on human-defined annotations. (2) Although adaptive semantic aggregation improves grounding and caption quality, we observe that the model may still generate coarse-grained captions for long events, motivating better information aggregation or representation learning mechanisms. (3) Our modified attention pattern is not fully compatible with some off-the-shelf optimized attention kernels, leaving room for custom kernel-level optimization to further improve efficiency.

\section*{Acknowledgments}
This research is supported by the National Research Foundation Singapore under its AI Singapore Programme (Award Number: AISG3-RP-2022-030). We would like to acknowledge that computational work involved in this research work is partially supported by NUS IT’s Research Computing group under grant number NUSREC-HPC-00001.

% ---- Main Bibliography ----
\putbib[main]
\end{bibunit}

\clearpage
\makeatletter
\def\@extra@binfo{.supp}
\def\@extra@b@citeb{.supp}
\makeatother
\begin{bibunit}[splncs04]
\title{Parallelized Autoregressive Decoding for Omni-Modal Dense Video Captioning} 

\titlerunning{PadCaptioner}

\author{Wenzheng Zeng, 
Siyi Jiao, 
Chen Gao, 
Hwee Tou Ng$^{\dagger}$, 
Mike Zheng Shou$^{\dagger}$} 

\authorrunning{W.~Zeng et al.}

\institute{National University of Singapore, Singapore \\
}

\maketitle
\begingroup
\def\thefootnote{}
\footnotetext[0]{$^{\dagger}$Corresponding authors.}
\endgroup

\appendix
\setcounter{figure}{0}
\setcounter{table}{0}
\setcounter{equation}{0}
\renewcommand{\thefigure}{A\arabic{figure}}
\renewcommand{\thetable}{A\arabic{table}}
\renewcommand{\theequation}{A\arabic{equation}}
\renewcommand{\theHsection}{appendix.\Alph{section}}
\renewcommand{\theHsubsection}{appendix.\Alph{section}.\arabic{subsection}}
\renewcommand{\theHfigure}{appendix.figure.\arabic{figure}}
\renewcommand{\theHtable}{appendix.table.\arabic{table}}
\renewcommand{\theHequation}{appendix.equation.\arabic{equation}}

% \begin{center}
%     \Large\bfseries Appendix
% \end{center}
{\centering\Large\bfseries Appendix\par}

\section{More Implementation Details}

Following~\cite{video_salmonn2}, we set the maximum number of pixels per frame to 61,250 for video frame inputs. Frames with higher resolution are resized proportionally to satisfy this constraint.

For the grounding constraint of the generated global event tokens, both the latent representation of the generated global event token $G^i$ and the video--audio prefix tokens are first projected through a lightweight two-layer MLP before computing feature similarity.
Unlike prior feature matching-based grounding methods~\cite{etbench_nips24, d2vlm_iccv25}, which operate under a simplified setting where each time segment (or frame) is represented by a single visual token and audio modality is ignored, we adopt a more general formulation that better aligns with modern multimodal LLMs.
Specifically, modern multimodal LLMs~\cite{qwen25omni,qwen25vl,qwen3_omni_arxiv25,video_salmonn2} typically encode the semantics of each time segment using multiple spatiotemporal tokens, and include tokens from both visual and audio modalities~\cite{qwen25omni,qwen3_omni_arxiv25,video_salmonn2}. 
Consequently, for each generated global event token $G^i$, we first compute its dot-product similarity with every video-audio prefix token and then aggregate the similarities of tokens belonging to the same segment by summation to obtain a segment-level similarity score, which is used to enforce the grounding constraint.

\begin{figure*}[t]
\footnotesize
\centering
\captionsetup{justification=raggedright,singlelinecheck=false}
\setlength{\tabcolsep}{4pt}

\begin{tabular*}{\textwidth}{@{}p{0.31\textwidth}@{\extracolsep{\fill}}p{0.31\textwidth}p{0.31\textwidth}@{}}
{\centering
\captionof{table}{Effect of similarity threshold on LongVALE.}
\label{tab:threshold_longvale}
\begin{tabular}{c|cc}
\toprule
Threshold & F1$\uparrow$ & Sim$\uparrow$ \\
\midrule
0.3 & 52.4 & 55.1 \\
0.4 & 52.9 & 55.4 \\
0.5 & 53.0 & 55.5 \\
0.6 & 52.4 & 55.5 \\
0.7 & 51.7 & 55.3 \\
\bottomrule
\end{tabular}
\par}
&
{\centering
\captionof{table}{Effect of similarity threshold on ChronusAV.}
\label{tab:threshold_chronusav}
\begin{tabular}{c|cc}
\toprule
Threshold & F1$\uparrow$ & Sim$\uparrow$ \\
\midrule
0.4 & 61.1 & 38.4 \\
0.5 & 61.4 & 38.5 \\
0.6 & 61.3 & 38.7 \\
0.7 & 61.8 & 38.8 \\
0.8 & 61.1 & 38.2 \\
\bottomrule
\end{tabular}
\par}
&
{\centering
\includegraphics[width=\linewidth,valign=t]{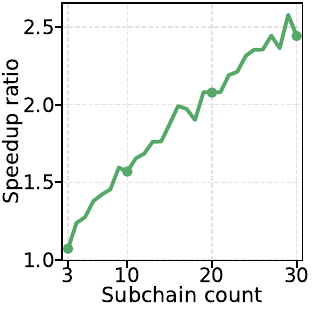}
\captionof{figure}{Speedup ratio over serial decoding across subchain count.}
\label{fig:speedup_analysis}
\par}
\end{tabular*}

\end{figure*}

In our framework, there are two ways to achieve event grounding during inference. 
(1) Grounding via textual timestamp generation. This is the most straightforward approach, as temporal information can be directly generated in textual form together with the captions.
(2) Grounding via feature similarity matching. 
As discussed in the main paper, the grounding constraints imposed during training endow each $G^i$ token with an intrinsic capability to distinguish the foreground event it represents from other segments. Consequently, event grounding can be performed at inference time by matching the generated global event token with the multimodal prefix tokens based on feature similarity.
Empirically, we find that the second approach yields better results. In practice, we adopt a similarity threshold to determine which temporal segments are salient (i.e., the grounding results). For each generated global event token, we compute its similarity with all video-audio prefix tokens using the same procedure as in the grounding constraint during training (i.e., token-level similarities are first aggregated within each time segment to obtain segment-level similarity scores). If the similarity score of a time segment exceeds a certain fraction of the maximum similarity across all segments for the current event token, the corresponding segment is regarded as salient and included in the grounded interval. We find that the choice of this threshold is fairly robust (e.g., an intuitive value of $0.5$ already achieves satisfactory performance, and the results remain stable within a reasonable range), as shown in~\cref{tab:threshold_longvale} and~\cref{tab:threshold_chronusav}. In practice, we set the threshold to $0.5$ for LongVALE~\cite{longvale} and $0.7$ for ChronusAV~\cite{chronusomni_arxiv25}.

Beyond dense video captioning (DVC), our DVC-oriented framework can generalize to a broader family of temporally grounded video understanding tasks, as demonstrated in the main paper. 
This generalization arises from the observation that dense video captioning inherently requires the joint modeling of temporal grounding and textual generation. The integration of these two capabilities naturally enables the framework to accommodate related tasks that emphasize either grounding-focused localization or textual response generation.

\begin{figure*}[t]
  \centering
  \includegraphics[width=0.98\linewidth]{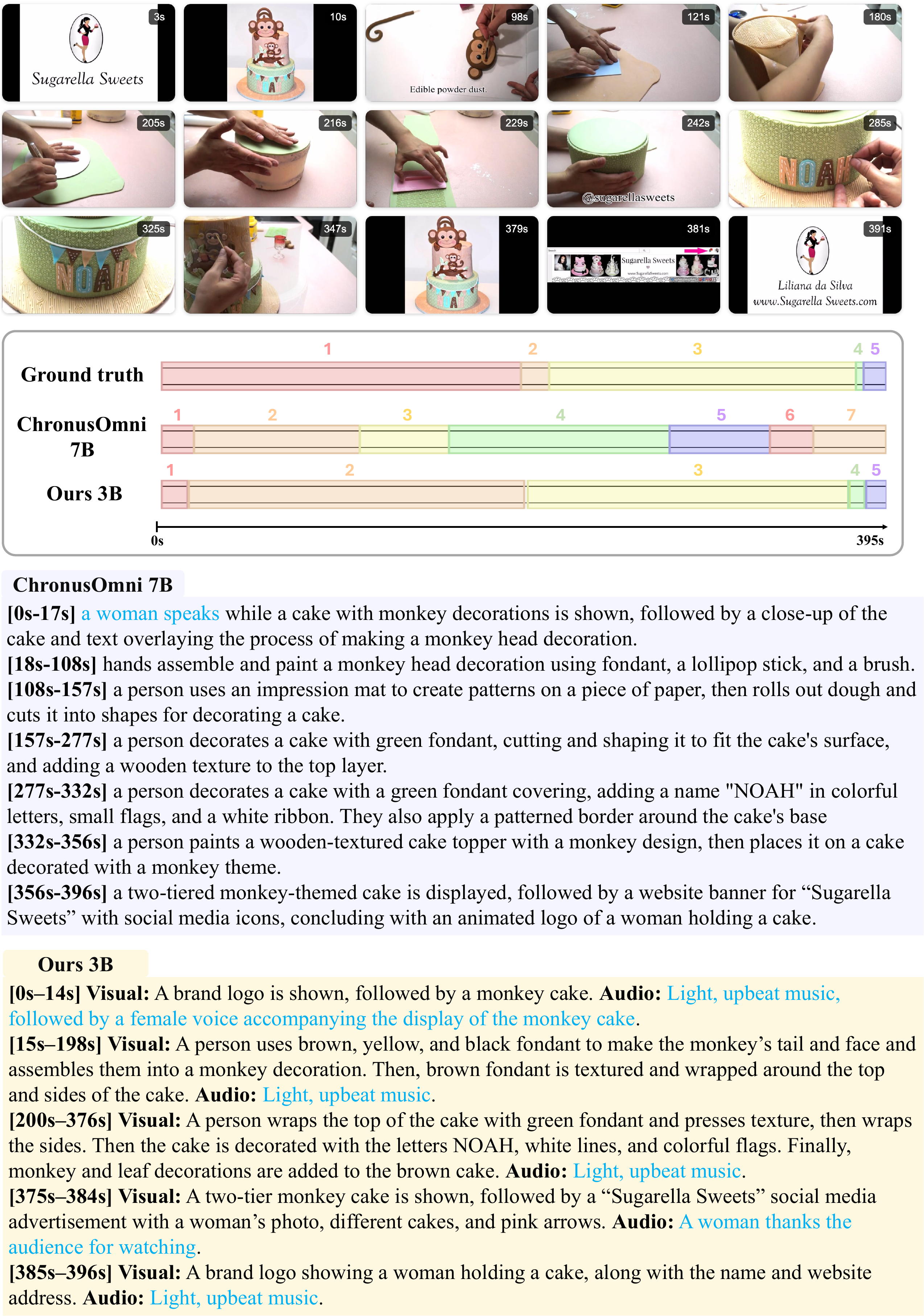}
    \caption{A qualitative example (YouTube ID: Za0fFT5NTbk) for the dense video captioning task. \textcolor[RGB]{0,176,240}{Blue} text indicates audio-related description.}
\label{fig:dvc}
\end{figure*}

\begin{figure*}[t]
  \centering
  \includegraphics[width=0.98\linewidth]{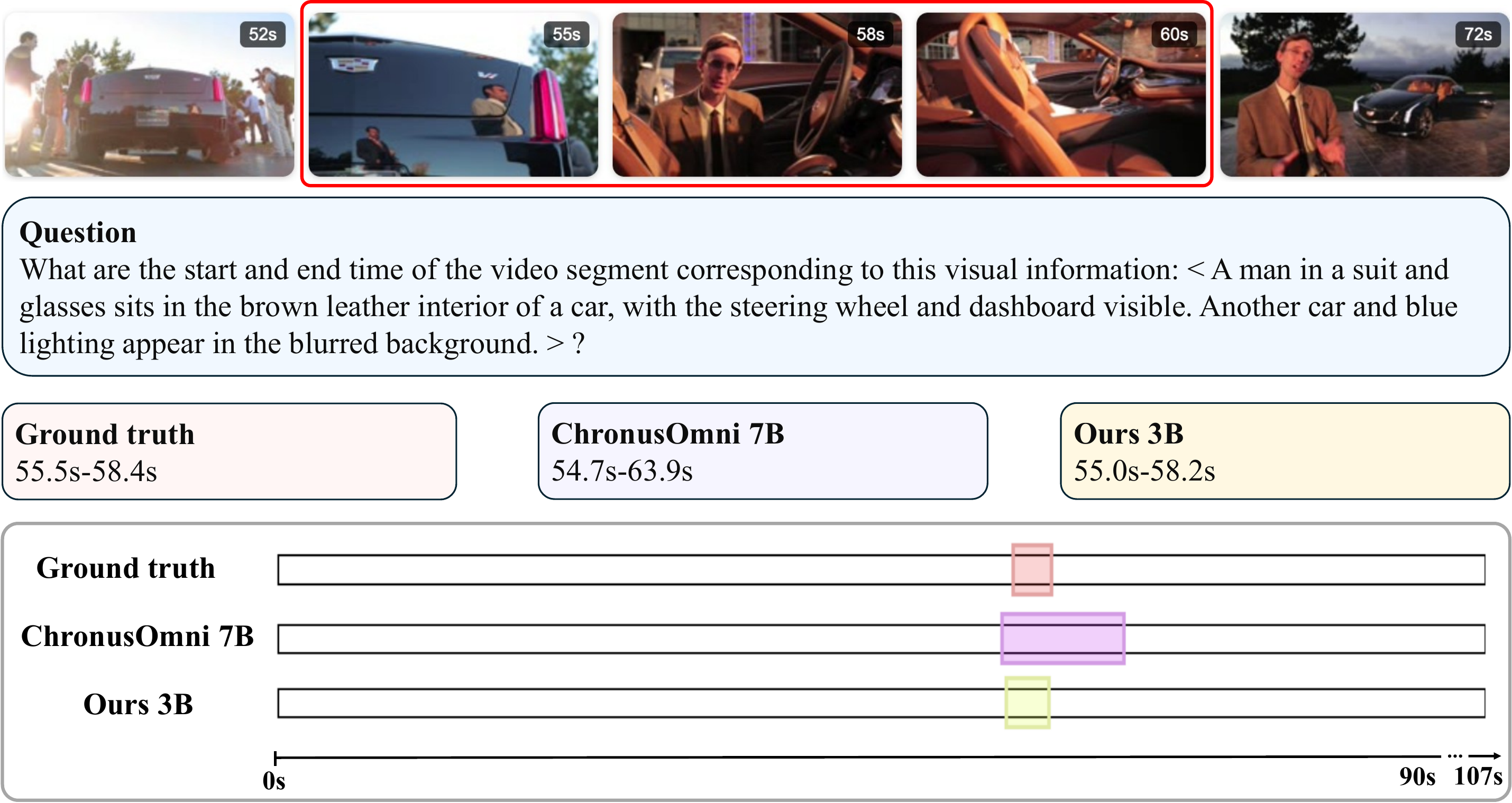}   
    \caption{A qualitative example (YouTube ID: pEkr5UJlrk8) for the V2T (video-to-time) task~\cite{chronusomni_arxiv25}.}
\label{fig:v2t}
\end{figure*}

\begin{figure*}[t]
  \centering
  \includegraphics[width=0.98\linewidth]{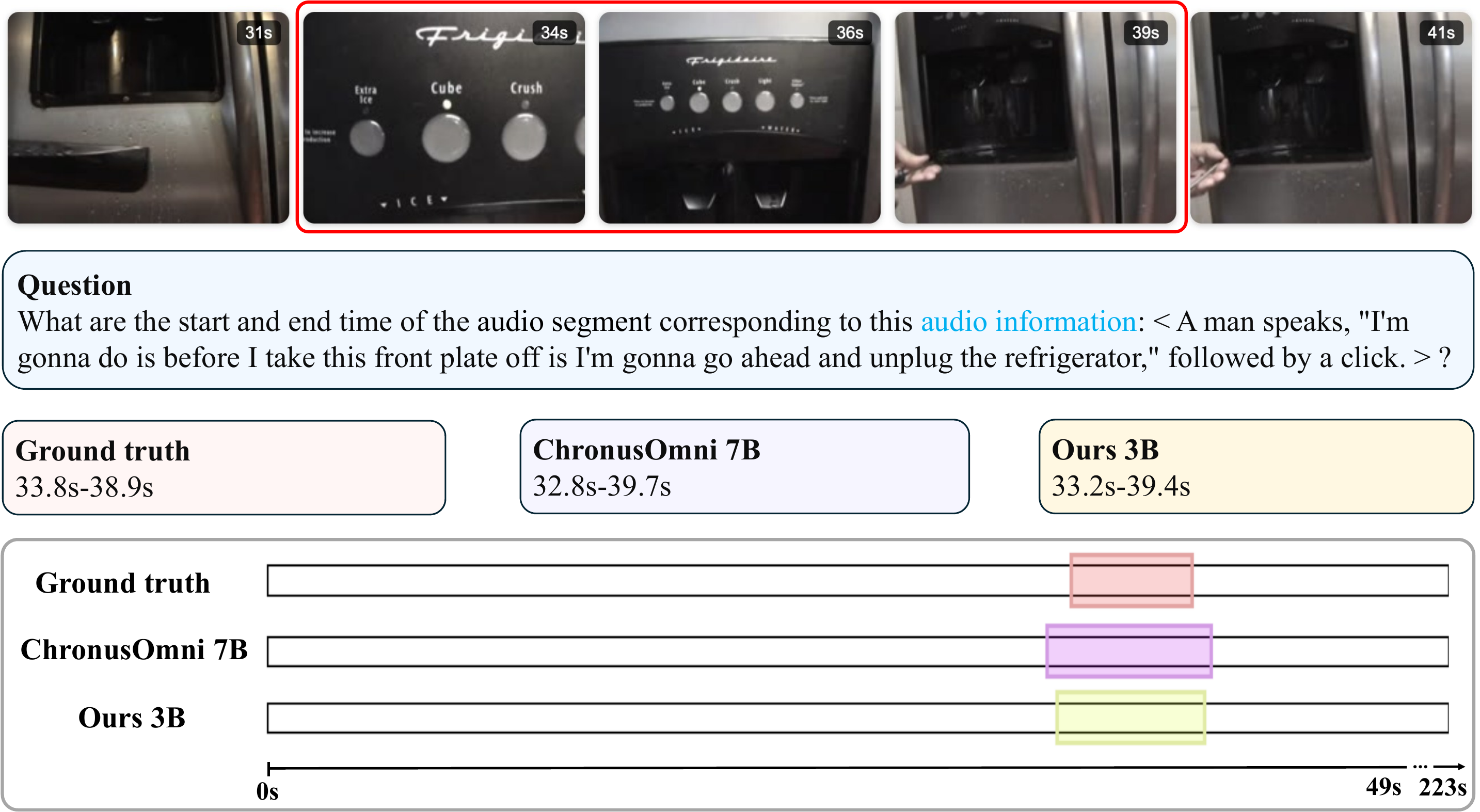} 
    \caption{A qualitative example (YouTube ID: 0Tmpt63T79M) for the A2T (audio-to-time) task~\cite{chronusomni_arxiv25}.}
\label{fig:a2t}
\end{figure*}

\begin{figure*}[t]
  \centering
  \includegraphics[width=0.98\linewidth]{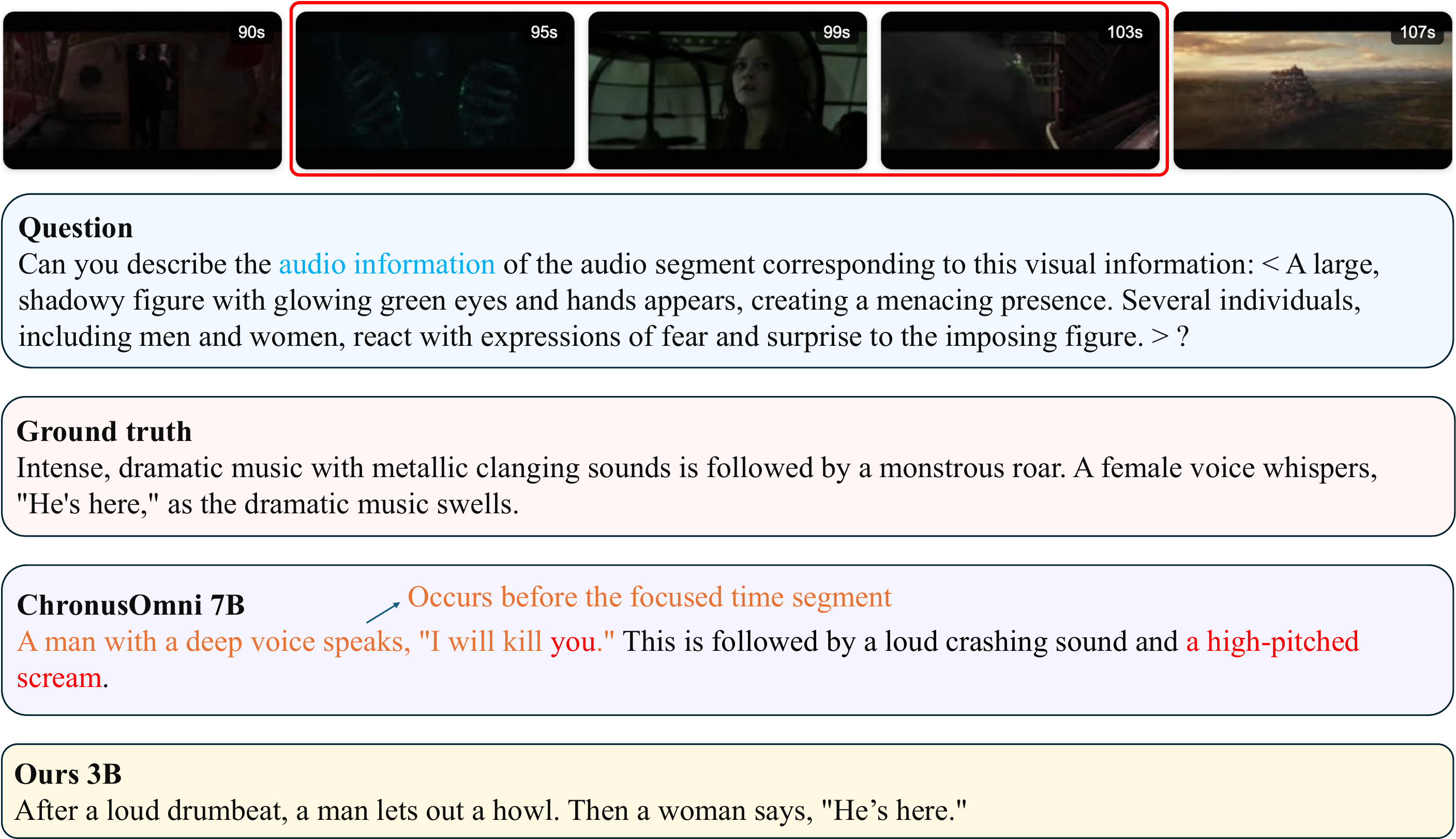} 
    \caption{A qualitative example (YouTube ID: i9jTOj4Bqo4) for the V2A (video-to-audio) task~\cite{chronusomni_arxiv25}. \textcolor{red}{Red} text indicates incorrect/hallucinated descriptions.}
\label{fig:v2a}
\end{figure*}

\begin{figure*}[t]
  \centering
  \includegraphics[width=0.98\linewidth]{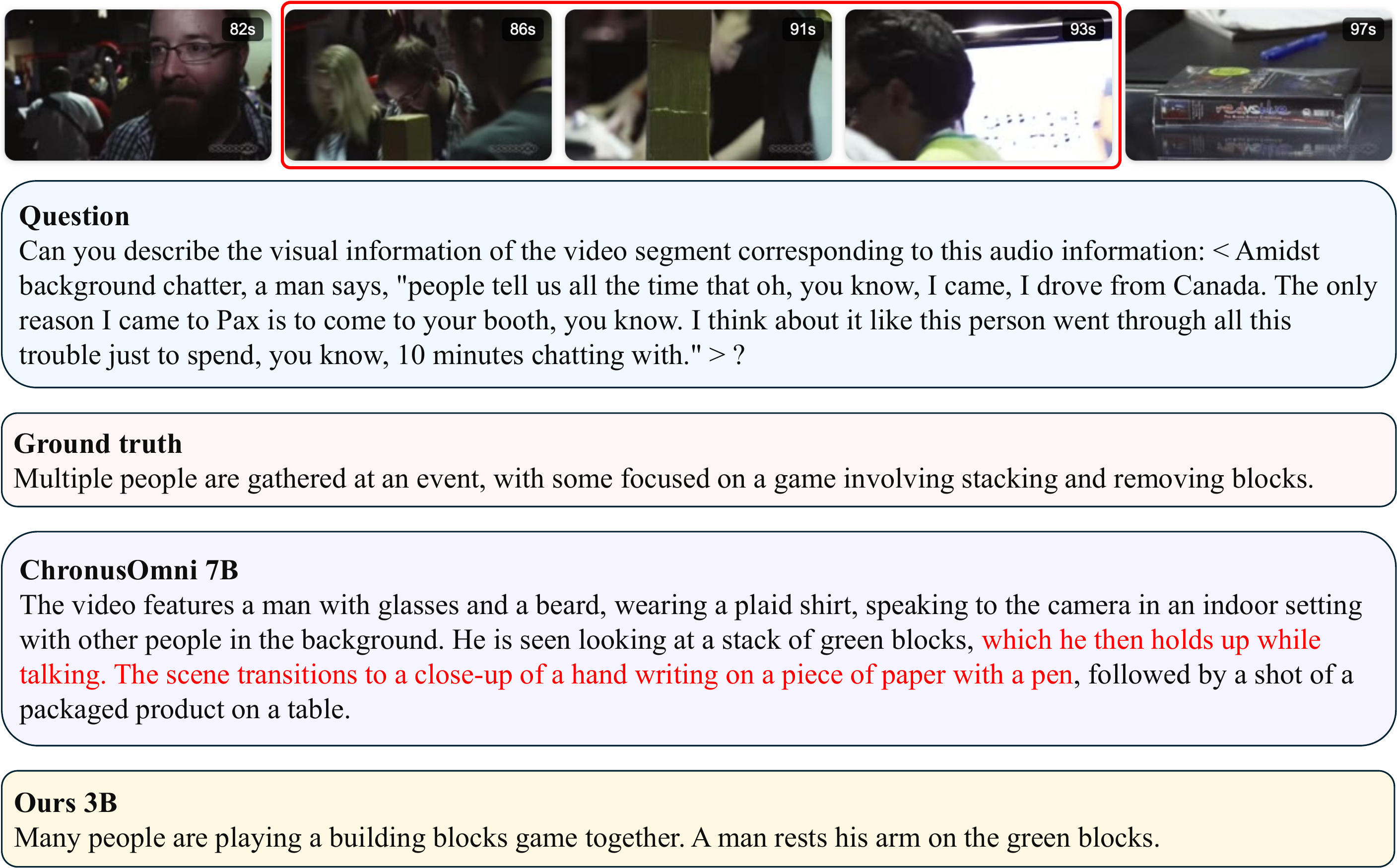}  
    \caption{A qualitative example (YouTube ID: EEGjQrxb9So) for the A2V (audio-to-video) task~\cite{chronusomni_arxiv25}. \textcolor{red}{Red} text indicates incorrect/hallucinated descriptions.}
\label{fig:a2v}
\end{figure*}

\begin{figure*}[t]
  \centering
  \includegraphics[width=0.98\linewidth]{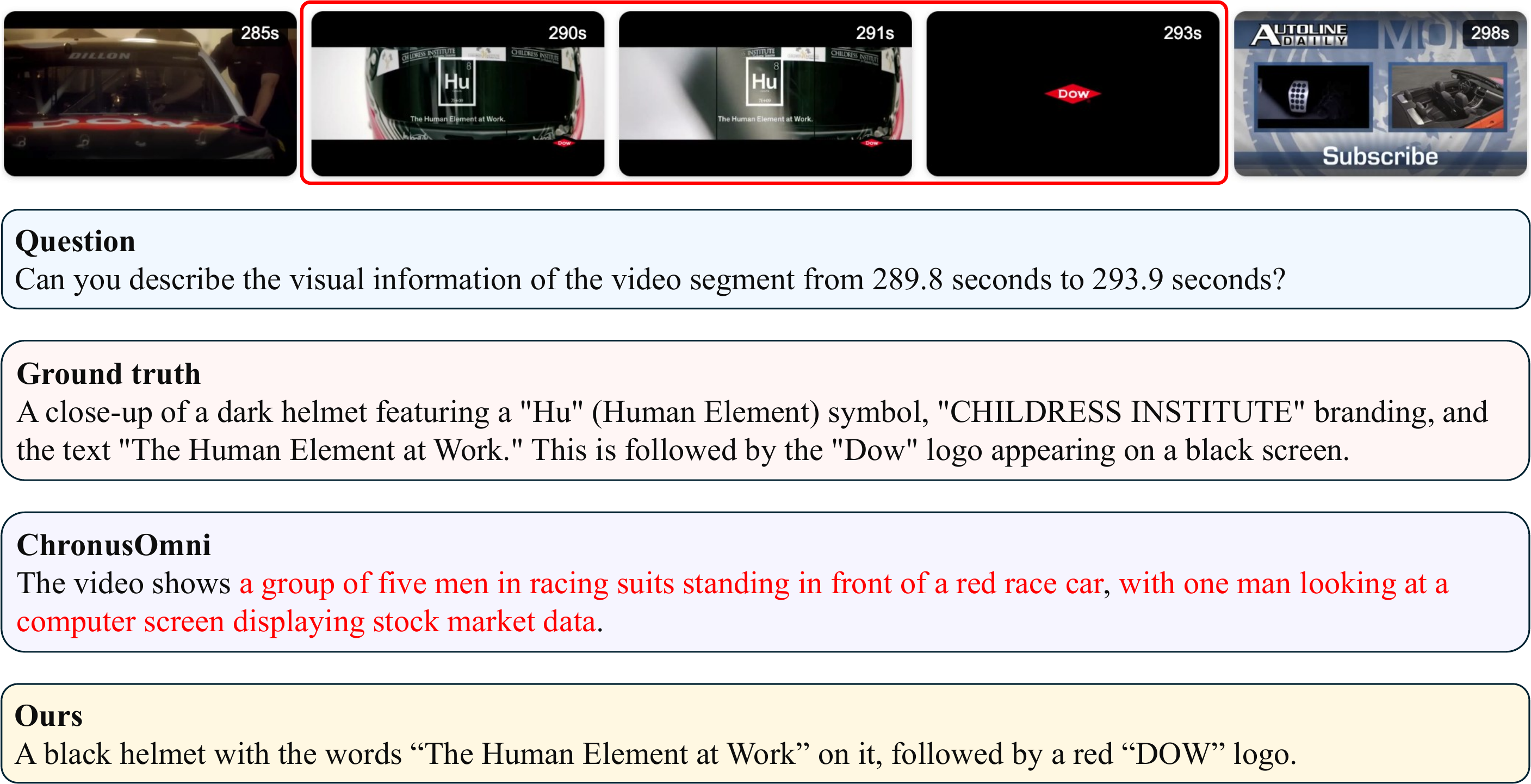}   
    \caption{A qualitative example (YouTube ID: d8okJ2zYyEc) for the T2V (time-to-video) task~\cite{chronusomni_arxiv25}. \textcolor{red}{Red} text indicates incorrect/hallucinated descriptions.}
\label{fig:t2v}
\end{figure*}

\begin{figure*}[t]
  \centering
  \includegraphics[width=0.98\linewidth]{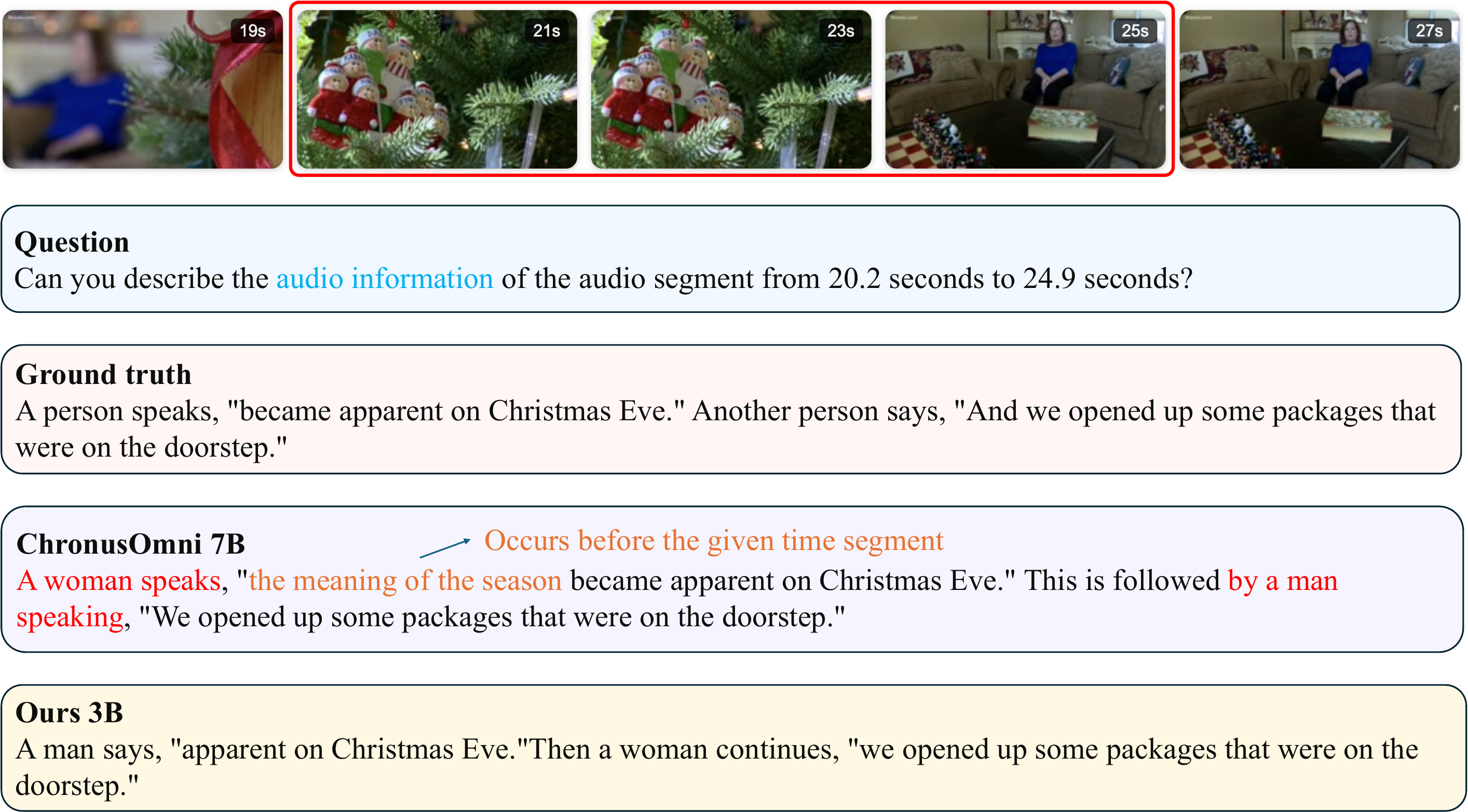}   
    \caption{A qualitative example (YouTube ID: JuS41kVKnF0) for the T2A (time-to-audio) task~\cite{chronusomni_arxiv25}. \textcolor{red}{Red} text indicates incorrect/hallucinated descriptions.}
\label{fig:t2a}
\end{figure*}

\begin{figure*}[t]
  \centering
  \includegraphics[width=0.98\linewidth]{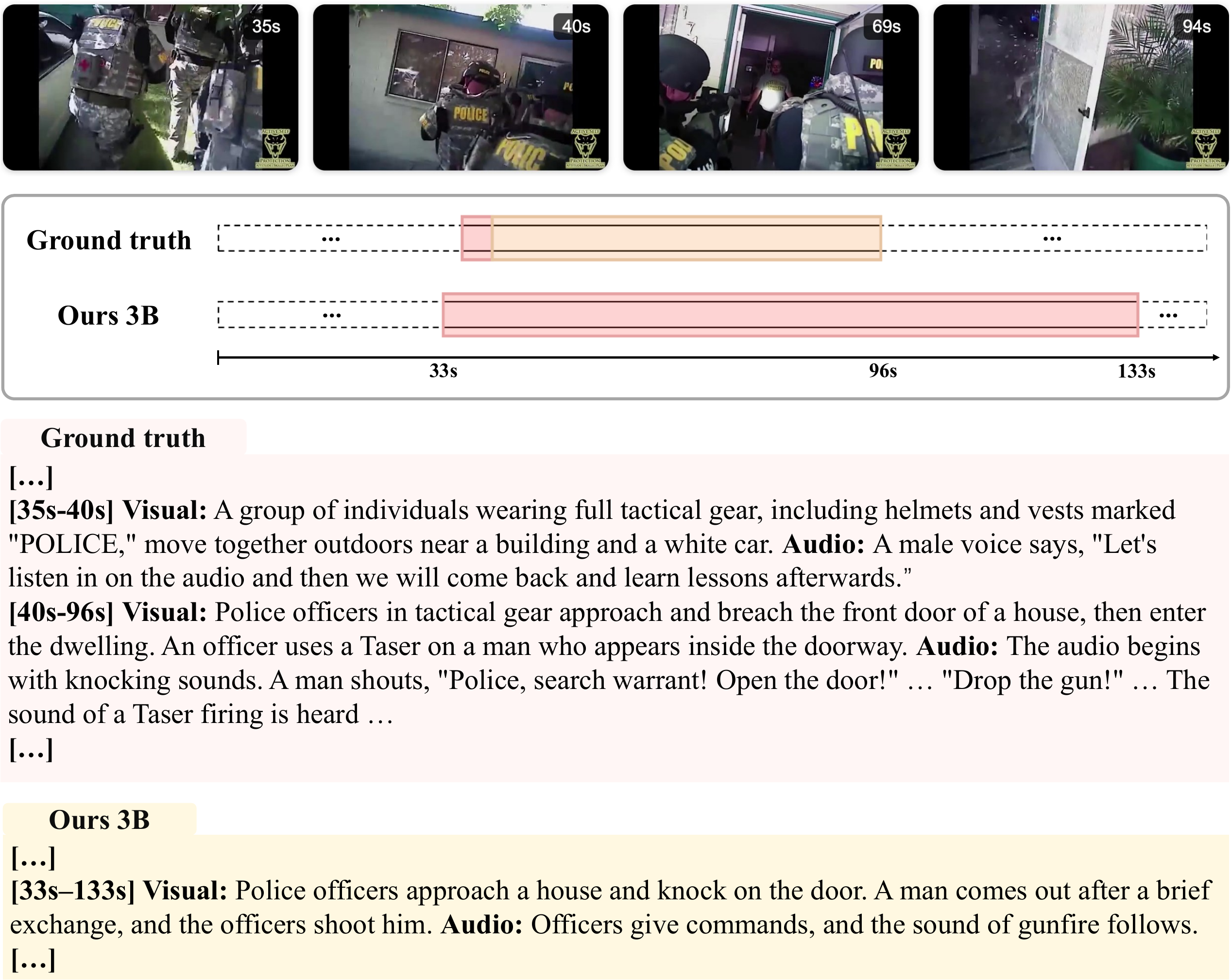}   
    \caption{Failure case analysis (YouTube ID: ektI4VWkU7o). We use “...” to omit part of the ASR transcript for brevity.}
\label{fig:failure}
\end{figure*}

\section{More Empirical Analysis}
\subsection{The Impact of Caption Length}
We find that longer captions do not necessarily imply better quality: 
(1) Naive base model fine-tuning (without our designs) increases the average caption length by $2.5\times$ without improving caption metrics; 
(2) Our method outperforms the previous SOTA~\cite{chronusomni_arxiv25} with shorter average caption length (549 vs. 685 tokens on ChronusAV~\cite{chronusomni_arxiv25}).

\begin{table}[t]
\centering
\caption{Base-model generalization on ChronusAV with fair 12K data training.}
\setlength{\tabcolsep}{8pt}
\begin{tabular}{lcc}
\toprule
Method & F1 & Sim \\
\midrule
LongVALE-LLM & 24.5 & 21.0 \\
LongVALE-LLM + Ours & \textbf{36.5} & \textbf{25.8} \\
\bottomrule
\end{tabular}
\label{tab:change_backbone}
\end{table}

\begin{table}[t]
\centering
\footnotesize
\caption{Training cost on 4 H200 GPUs (bs: batch size).}
\setlength{\tabcolsep}{8pt}
\begin{tabular}{lcc}
\toprule
Method & Memory (bs) & Time (bs per GPU) \\
\midrule
Ours (without FlashAttention-2) & 78GB (1) & 28h (2) \\
Standard (with FlashAttention-2) & 58GB (1) & 20h or 15h (2 or 3) \\
\bottomrule
\end{tabular}
\label{tab:training_cost}
\end{table}

\subsection{Base-Model Generalization}

Beyond the main base model adopted in the main paper~\cite{video_salmonn2}, we further examine whether our proposed components generalize to a different base model~\cite{longvale}. The result in Tab.~\ref{tab:change_backbone} demonstrates consistent improvements, suggesting that our method is not tied to a specific base model.

\subsection{Degree of Parallelization}
It is driven by the number of model-predicted event subchains (13.2 on average). 
Speedup over serial decoding increases with the subchain count, following a log-like trend (Fig.~\ref{fig:speedup_analysis}), while parallel decoding does not degrade performance, as validated in the main paper.

\subsection{Training Cost}
We analyze the training cost under a 12K-sample training budget. Our method requires modifying standard causal attention, which is not fully compatible with some off-the-shelf optimized attention kernels (e.g., FlashAttention-2~\cite{flashattention2}). This is especially the case because our current implementation does not include dedicated kernel-level optimization. As shown in Tab.~\ref{tab:training_cost}, without an optimized kernel implementation, our method incurs about $0.3\times$ extra memory usage and $1.4\times$/$1.9\times$ longer training time under the same/maximum feasible batch size on H200 GPUs. However, after being trained on only 12K samples, our method provides clear inference-time advantages, and its efficiency can be further improved with custom kernel optimization.

\subsection{Inference Requirement}
Compared with standard autoregressive decoding, parallelized decoding itself does not increase the KV-cache size (assuming that the total generated caption length remains the same). The additional peak GPU memory mainly comes from transient activations when predicting $K$ tokens simultaneously at each decoding step (rather than a single token), and thus remains manageable. The maximum GPU memory usage is below 30GB.

\section{Qualitative Analysis}

We present qualitative analyses to better demonstrate the effectiveness of our approach. Beyond the dense video captioning task, we further analyze the performance of our method on other temporally grounded video understanding scenarios. These analyses aim to validate the generalization ability of our framework and provide a deeper understanding of its capabilities from multiple perspectives. We also include predictions from the recent state-of-the-art model ChronusOmni~\cite{chronusomni_arxiv25} for comparison.
Finally, we provide a failure case analysis to reveal the limitations of our current framework and motivate future research.

% \subsection{Dense video captioning task}
\smallskip
\noindent \textbf{Dense video captioning.} 
From~\cref{fig:dvc}, several observations can be made. 
(1) Compared with the recent SOTA, our method achieves more reasonable and accurate temporal localization of events, producing boundaries that better align with the actual semantic transitions in the video. 
(2) Our model also generates more coherent and informative captions, particularly in terms of audio-aware perception. While the compared method is also equipped with audio perception capability, it often fails to effectively exploit acoustic cues, whereas our approach correctly captures audio-related events such as background music and speech. In addition, our model can produce structured and disentangled visually oriented and audio-oriented descriptions, which can benefit downstream applications such as modality-aware retrieval or multimodal content indexing.
(3) These advantages are achieved with a relatively small model size (3B) and significant speedup compared with the SOTA method~\cite{chronusomni_arxiv25}, as verified in the main paper. This further highlights the effectiveness of our proposed approach.

\smallskip
\noindent \textbf{Omni temporal grounding.} 
We further evaluate the generalization ability of our model on the omni temporal grounding task. Specifically, we consider two settings following the task definition in~\cite{chronusomni_arxiv25}: video-to-time (V2T) and audio-to-time (A2T) grounding, which aim to respectively localize the temporal interval in the video that matches the visual or audio information described in the user query. 
From~\cref{fig:v2t} and~\cref{fig:a2t}, it can be observed that our method produces more accurate temporal boundaries, while the strong baseline tends to include additional irrelevant segments. 
Notably, in the example shown in~\cref{fig:a2t}, the visual appearance around the target event is highly similar across neighboring frames, making it difficult to localize the event based solely on visual cues. This scenario requires the model to rely on audio information for accurate grounding. Our method successfully identifies the correct temporal interval under such conditions, demonstrating its stronger audio-aware temporal localization capability.

\smallskip
\noindent \textbf{Cross-modal grounding and captioning.}
We also evaluate our model on the A2V and V2A tasks defined in~\cite{chronusomni_arxiv25}, which require compositional reasoning across modalities: the model must first ground the event of interest based on the description from one modality, and then describe the grounded temporal segment from the complementary modality. Specifically, audio-to-video (A2V) requires localizing the event according to the audio query and generating its visual description, while video-to-audio (V2A) requires grounding the event from visual cues and then characterizing it from the audio perspective. From~\cref{fig:v2a} and~\cref{fig:a2v}, it can be observed that the recent strong counterpart frequently produces hallucinated descriptions that refer to events not present in the grounded video segment or even in the entire video. In contrast, our method correctly reasons about the corresponding temporal segment and generates faithful descriptions without hallucination.

\smallskip
\noindent \textbf{Omni segmentation captioning.}
In this setting, the model is required to generate a description for a user-specified temporal interval. This task evaluates the model’s ability to correctly interpret the provided time span and to understand and describe the corresponding event. Depending on the target modality of the description, the task can be further divided into two variants: time-to-video (T2V), which requires generating a visual description of the specified segment, and time-to-audio (T2A), which requires describing the audio content within the given temporal interval~\cite{chronusomni_arxiv25}. 
As shown in~\cref{fig:t2v} and~\cref{fig:t2a}, the recent strong baseline often produces descriptions that are not faithfully grounded in the specified temporal segment (\cref{fig:t2v}). In the T2A example (\cref{fig:t2a}), the counterpart~\cite{chronusomni_arxiv25} fails to distinguish the speaker’s gender and tends to include information beyond the specified temporal interval. In contrast, our method generates descriptions that remain aligned with the given time span, demonstrating stronger fine-grained audio understanding and more precise temporal grounding despite using a relatively lightweight model.

\smallskip
\noindent \textbf{Failure case analysis.}
Despite its effectiveness, our method still has some notable limitations. First, event grounding is still imperfect, and grounding errors can propagate to caption generation. 
For example, as shown in~\cref{fig:failure}, our model fails to split adjacent events. This is mainly due to its limited ability to distinguish fine-grained semantic differences between temporally close and visually similar events.
In addition, from~\cref{fig:failure}, we can also observe that our method captures the key content of the event (i.e., police officers approach a house and confront a man inside). However, the generated caption still lacks fine-grained details, especially in faithfully preserving speech content. For instance, it only briefly mentions that the officers give commands, without preserving the specific spoken content. This limitation motivates future research on more effective information aggregation or representation learning for long and information-dense events.

\renewcommand{\refname}{Appendix References}
\putbib[main]
\end{bibunit}


\begin{thebibliography}{10}
\providecommand{\url}[1]{\texttt{#1}}
\providecommand{\urlprefix}{URL }
\providecommand{\doi}[1]{https://doi.org/#1}

\bibitem{block_diffusion_iclr25}
Arriola, M., Gokaslan, A., Chiu, J.T., Yang, Z., Qi, Z., Han, J., Sahoo, S.S., Kuleshov, V.: Block diffusion: Interpolating between autoregressive and diffusion language models. In: The Thirteenth International Conference on Learning Representations (2025)

\bibitem{qwen3vl_arxiv25}
Bai, S., Cai, Y., Chen, R., Chen, K., Chen, X., Cheng, Z., Deng, L., Ding, W., Gao, C., Ge, C., et~al.: Qwen3-vl technical report. arXiv preprint arXiv:2511.21631  (2025)

\bibitem{qwen25vl}
Bai, S., Chen, K., Liu, X., Wang, J., Ge, W., Song, S., Dang, K., Wang, P., Wang, S., Tang, J., et~al.: Qwen2. 5-vl technical report. arXiv preprint arXiv:2502.13923  (2025)

\bibitem{auroracap_iclr25}
Chai, W., Song, E., Du, Y., Meng, C., Madhavan, V., Bar-Tal, O., Hwang, J.N., Xie, S., Manning, C.D.: Auroracap: Efficient, performant video detailed captioning and a new benchmark. In: The Thirteenth International Conference on Learning Representations (2025)

\bibitem{videollm_online_cvpr24}
Chen, J., Lv, Z., Wu, S., Lin, K.Q., Song, C., Gao, D., Liu, J.W., Gao, Z., Mao, D., Shou, M.Z.: Videollm-online: Online video large language model for streaming video. In: Proceedings of the IEEE/CVF Conference on Computer Vision and Pattern Recognition. pp. 18407--18418 (2024)

\bibitem{livecc_cvpr25}
Chen, J., Zeng, Z., Lin, Y., Li, W., Ma, Z., Shou, M.Z.: Livecc: Learning video llm with streaming speech transcription at scale. In: Proceedings of the Computer Vision and Pattern Recognition Conference. pp. 29083--29095 (2025)

\bibitem{avocado_video_captioning_iclr26}
Chen, X., Ding, Y., Lin, W., Hua, J., Yao, L., Shi, Y., Li, B., Zhang, Y., Liu, Q., Wan, P., et~al.: Avocado: An audiovisual video captioner driven by temporal orchestration. In: International Conference on Learning Representations (2026)

\bibitem{chronusomni_arxiv25}
Chen, Y., Wu, Y., Guan, K., Ren, Y., Wang, Y., Song, R., Ru, L.: Chronusomni: Improving time awareness of omni large language models. arXiv preprint arXiv:2512.09841  (2025)

\bibitem{videotg_r1_arxiv25}
Dong, L., Zhang, H., Lin, H., Yan, Z., Zeng, X., Zhang, H., Huang, Y., Wang, Y., Ling, Z.H., Wang, L., et~al.: Videotg-r1: Boosting video temporal grounding via curriculum reinforcement learning on reflected boundary annotations. arXiv preprint arXiv:2510.23397  (2025)

\bibitem{interleaved_latent_reasoning_arxiv25}
Dong, S., Wang, S., Liu, X., Li, C., Hou, H., Wei, Z.: Interleaved latent visual reasoning with selective perceptual modeling. arXiv preprint arXiv:2512.05665  (2025)

\bibitem{palm_e_embodied_vlm_agent_dvc_app_icml24}
Driess, D., Xia, F., Sajjadi, M.S., Lynch, C., Chowdhery, A., Ichter, B., Wahid, A., Tompson, J., Vuong, Q., Yu, T., et~al.: Palm-e: an embodied multimodal language model. In: Proceedings of the 40th International Conference on Machine Learning. pp. 8469--8488 (2023)

\bibitem{arc_hunyuan_video_arxiv25}
Ge, Y., Ge, Y., Li, C., Wang, T., Pu, J., Li, Y., Qiu, L., Ma, J., Duan, L., Zuo, X., et~al.: Arc-hunyuan-video-7b: Structured video comprehension of real-world shorts. arXiv preprint arXiv:2507.20939  (2025)

\bibitem{longvale}
Geng, T., Zhang, J., Wang, Q., Wang, T., Duan, J., Zheng, F.: Longvale: Vision-audio-language-event benchmark towards time-aware omni-modal perception of long videos. In: Proceedings of the Computer Vision and Pattern Recognition Conference. pp. 18959--18969 (2025)

\bibitem{ego4d_cvpr22}
Grauman, K., Westbury, A., Byrne, E., Chavis, Z., Furnari, A., Girdhar, R., Hamburger, J., Jiang, H., Liu, M., Liu, X., et~al.: Ego4d: Around the world in 3,000 hours of egocentric video. In: Proceedings of the IEEE/CVF conference on computer vision and pattern recognition. pp. 18995--19012 (2022)

\bibitem{ego_exo_4d_cvpr24}
Grauman, K., Westbury, A., Torresani, L., Kitani, K., Malik, J., Afouras, T., Ashutosh, K., Baiyya, V., Bansal, S., Boote, B., et~al.: Ego-exo4d: Understanding skilled human activity from first-and third-person perspectives. In: Proceedings of the IEEE/CVF Conference on Computer Vision and Pattern Recognition. pp. 19383--19400 (2024)

\bibitem{vtgllm_aaai25}
Guo, Y., Liu, J., Li, M., Cheng, D., Tang, X., Sui, D., Liu, Q., Chen, X., Zhao, K.: Vtg-llm: Integrating timestamp knowledge into video llms for enhanced video temporal grounding. arXiv preprint arXiv:2405.13382  (2024)

\bibitem{trace_iclr25}
Guo, Y., Liu, J., Li, M., Tang, X., Liu, Q., Chen, X.: Trace: Temporal grounding video llm via causal event modeling. In: International Conference on Learning Representations (2025)

\bibitem{nar_iccv25}
He, Y., He, Y., He, S., Chen, F., Zhou, H., Zhang, K., Zhuang, B.: Neighboring autoregressive modeling for efficient visual generation. In: Proceedings of the IEEE/CVF International Conference on Computer Vision. pp. 19000--19010 (2025)

\bibitem{vidlada_video_dllm_arxiv26}
He, Z., Chen, T., Wang, K., Qin, Z., Shao, Y., Gan, C., Li, S., Wu, Z., Lin, W.: Vidlada: Bidirectional diffusion large language models for efficient video understanding. arXiv preprint arXiv:2601.17868  (2026)

\bibitem{lora}
Hu, E.J., Shen, Y., Wallis, P., Allen-Zhu, Z., Li, Y., Wang, S., Wang, L., Chen, W., et~al.: Lora: Low-rank adaptation of large language models. ICLR  \textbf{1}(2), ~3 (2022)

\bibitem{vtimellm}
Huang, B., Wang, X., Chen, H., Song, Z., Zhu, W.: Vtimellm: Empower llm to grasp video moments. In: Proceedings of the IEEE/CVF Conference on Computer Vision and Pattern Recognition. pp. 14271--14280 (2024)

\bibitem{dvc_query_based_cvpr24}
Kim, M., Kim, H.B., Moon, J., Choi, J., Kim, S.T.: Do you remember? dense video captioning with cross-modal memory retrieval. In: Proceedings of the IEEE/CVF Conference on Computer Vision and Pattern Recognition. pp. 13894--13904 (2024)

\bibitem{hicm2_aaai25}
Kim, M., Kim, H.B., Moon, J., Choi, J., Kim, S.T.: Hicm$^2$: Hierarchical compact memory modeling for dense video captioning. In: Proceedings of the AAAI Conference on Artificial Intelligence. pp. 4293--4301 (2025)

\bibitem{activitynet_caption_iccv17}
Krishna, R., Hata, K., Ren, F., Fei-Fei, L., Carlos~Niebles, J.: Dense-captioning events in videos. In: Proceedings of the IEEE international conference on computer vision. pp. 706--715 (2017)

\bibitem{zebra_cot_iclr26}
Li, A., Wang, C., Fu, D., Yue, K., Cai, Z., Zhu, W.B., Liu, O., Guo, P., Neiswanger, W., Huang, F., et~al.: Zebra-cot: A dataset for interleaved vision language reasoning. In: International Conference on Learning Representations (2026)

\bibitem{ARPG_iclr26}
Li, H., Yang, J., Li, G., Wang, H.: Autoregressive image generation with randomized parallel decoding. In: The Fourteenth International Conference on Learning Representations (2026)

\bibitem{video_opd_temporal_grounding_arxiv26}
Li, J., Yin, H., Xu, H., Xu, B., Tan, W., He, Z., Ju, J., Luo, Z., Luan, J.: Video-opd: Efficient post-training of multimodal large language models for temporal video grounding via on-policy distillation. arXiv preprint arXiv:2602.02994  (2026)

\bibitem{videochat}
Li, K., He, Y., Wang, Y., Li, Y., Wang, W., Luo, P., Wang, Y., Wang, L., Qiao, Y.: Videochat: Chat-centric video understanding. Science China Information Sciences  \textbf{68}(10),  200102 (2025)

\bibitem{videochat_flash}
Li, X., Wang, Y., Yu, J., Zeng, X., Zhu, Y., Huang, H., Gao, J., Li, K., He, Y., Wang, C., et~al.: Videochat-flash: Hierarchical compression for long-context video modeling. arXiv preprint arXiv:2501.00574  (2025)

\bibitem{videochat_r1_arxiv25}
Li, X., Yan, Z., Meng, D., Dong, L., Zeng, X., He, Y., Wang, Y., Qiao, Y., Wang, Y., Wang, L.: Videochat-r1: Enhancing spatio-temporal perception via reinforcement fine-tuning. arXiv preprint arXiv:2504.06958  (2025)

\bibitem{jointly_dvc_cvpr18}
Li, Y., Yao, T., Pan, Y., Chao, H., Mei, T.: Jointly localizing and describing events for dense video captioning. In: Proceedings of the IEEE conference on computer vision and pattern recognition. pp. 7492--7500 (2018)

\bibitem{lira_seg_iccv25}
Li, Z., Yang, B., Liu, Q., Zhang, S., Ma, Z., Yin, L., Deng, L., Sun, Y., Liu, Y., Bai, X.: Lira: Inferring segmentation in large multi-modal models with local interleaved region assistance. In: Proceedings of the IEEE/CVF International Conference on Computer Vision. pp. 24056--24067 (2025)

\bibitem{trisense_nips25}
Li, Z., Zhang, X., Guo, Y., Bennamoun, M., Boussaid, F., Dwivedi, G., Gong, L., Ke, Q.: Watch and listen: Understanding audio-visual-speech moments with multimodal llm. In: The Thirty-ninth Annual Conference on Neural Information Processing Systems (2025)

\bibitem{etbench_nips24}
Liu, Y., Ma, Z., Qi, Z., Wu, Y., Shan, Y., Chen, C.W.: {E.T. Bench}: Towards open-ended event-level video-language understanding. In: The Thirty-eighth Conference on Neural Information Processing Systems Datasets and Benchmarks Track (2024)

\bibitem{dvc_query_based_acl25}
Liu, Z., Zhang, X., Liu, J.: Task-specific information decomposition for end-to-end dense video captioning. In: Proceedings of the 63rd Annual Meeting of the Association for Computational Linguistics (Volume 1: Long Papers). pp. 16524--16536 (2025)

\bibitem{ola_arxiv25}
Liu, Z., Dong, Y., Wang, J., Liu, Z., Hu, W., Lu, J., Rao, Y.: Ola: Pushing the frontiers of omni-modal language model. arXiv preprint arXiv:2502.04328  (2025)

\bibitem{video_chatgpt_acl24}
Maaz, M., Rasheed, H., Khan, S., Khan, F.: Video-chatgpt: Towards detailed video understanding via large vision and language models. In: Proceedings of the 62nd Annual Meeting of the Association for Computational Linguistics (Volume 1: Long Papers). pp. 12585--12602 (2024)

\bibitem{streamlined_dvc_two_stage_cvpr19}
Mun, J., Yang, L., Ren, Z., Xu, N., Han, B.: Streamlined dense video captioning. In: Proceedings of the IEEE/CVF conference on computer vision and pattern recognition. pp. 6588--6597 (2019)

\bibitem{llada_arxiv25}
Nie, S., Zhu, F., You, Z., Zhang, X., Ou, J., Hu, J., Zhou, J., Lin, Y., Wen, J.R., Li, C.: Large language diffusion models. arXiv preprint arXiv:2502.09992  (2025)

\bibitem{randar_cvpr25}
Pang, Z., Zhang, T., Luan, F., Man, Y., Tan, H., Zhang, K., Freeman, W.T., Wang, Y.X.: Randar: Decoder-only autoregressive visual generation in random orders. In: Proceedings of the Computer Vision and Pattern Recognition Conference. pp. 45--55 (2025)

\bibitem{meco_iclr26}
Pang, Z., Otani, M., Nakashima, Y.: Measure twice, cut once: Grasping video structures and event semantics with llms for video temporal localization. In: International Conference on Learning Representations (2026)

\bibitem{arc_chapter_arxiv25}
Pu, J., Wang, T., Ge, Y., Ge, Y., Li, C., Shan, Y.: Arc-chapter: Structuring hour-long videos into navigable chapters and hierarchical summaries. arXiv preprint arXiv:2511.14349  (2025)

\bibitem{momentor_icml2024}
Qian, L., Li, J., Wu, Y., Ye, Y., Fei, H., Chua, T.S., Zhuang, Y., Tang, S.: Momentor: Advancing video large language model with fine-grained temporal reasoning. arXiv preprint arXiv:2402.11435  (2024)

\bibitem{d3llm_arxiv26}
Qian, Y.Y., Su, J., Hu, L., Zhang, P., Deng, Z., Zhao, P., Zhang, H.: d3llm: Ultra-fast diffusion llm using pseudo-trajectory distillation. arXiv preprint arXiv:2601.07568  (2026)

\bibitem{whisper_v2_speech_encoder}
Radford, A., Kim, J.W., Xu, T., Brockman, G., McLeavey, C., Sutskever, I.: Robust speech recognition via large-scale weak supervision. In: International conference on machine learning. pp. 28492--28518. PMLR (2023)

\bibitem{timechat_cvpr24}
Ren, S., Yao, L., Li, S., Sun, X., Hou, L.: Timechat: A time-sensitive multimodal large language model for long video understanding. In: Proceedings of the IEEE/CVF Conference on Computer Vision and Pattern Recognition. pp. 14313--14323 (2024)

\bibitem{growing_a_twig_iccv25}
Shao, Z., Wang, M., Yu, Z., Pan, W., Yang, Y., Wei, T., Zhang, H., Mao, N., Chen, W., Yu, J.: Growing a twig to accelerate large vision-language models. In: Proceedings of the IEEE/CVF International Conference on Computer Vision. pp. 20064--20074 (2025)

\bibitem{pandagpt}
Su, Y., Lan, T., Li, H., Xu, J., Wang, Y., Cai, D.: Pandagpt: One model to instruction-follow them all. In: Proceedings of the 1st Workshop on Taming Large Language Models: Controllability in the era of Interactive Assistants! pp. 11--23 (2023)

\bibitem{padt_iclr26}
Su, Y., Zhang, H., Li, S., Liu, N., Liao, J., Pan, J., Liu, Y., Xing, X., Sun, C., Li, C., et~al.: Patch-as-decodable-token: Towards unified multi-modal vision tasks in mllms. In: International Conference on Learning Representations (2026)

\bibitem{video_salmonn2}
Tang, C., Li, Y., Yang, Y., Zhuang, J., Sun, G., Li, W., Ma, Z., Zhang, C.: video-salmonn 2: Captioning-enhanced audio-visual large language models. arXiv preprint arXiv:2506.15220  (2025)

\bibitem{audio_qformer_icassp24}
Tang, C., Yu, W., Sun, G., Chen, X., Tan, T., Li, W., Lu, L., Ma, Z., Zhang, C.: Extending large language models for speech and audio captioning. In: ICASSP 2024-2024 IEEE International Conference on Acoustics, Speech and Signal Processing (ICASSP). pp. 11236--11240. IEEE (2024)

\bibitem{avicuna_aaai25}
Tang, Y., Shimada, D., Bi, J., Feng, M., Hua, H., Xu, C.: Empowering llms with pseudo-untrimmed videos for audio-visual temporal understanding. In: Proceedings of the AAAI Conference on Artificial Intelligence. pp. 7293--7301 (2025)

\bibitem{sketch_in_latents_arxiv25}
Tong, J., Gu, J., Lou, Y., Fan, L., Zou, Y., Wu, Y., Ye, J., Li, R.: Sketch-in-latents: Eliciting unified reasoning in mllms. arXiv preprint arXiv:2512.16584  (2025)

\bibitem{pdvc_iccv21}
Wang, T., Zhang, R., Lu, Z., Zheng, F., Cheng, R., Luo, P.: End-to-end dense video captioning with parallel decoding. In: Proceedings of the IEEE/CVF international conference on computer vision. pp. 6847--6857 (2021)

\bibitem{d2f_dllm_arxiv25}
Wang, X., Xu, C., Jin, Y., Jin, J., Zhang, H., Deng, Z.: Diffusion llms can do faster-than-ar inference via discrete diffusion forcing. arXiv preprint arXiv:2508.09192  (2025)

\bibitem{timer1_arxiv25}
Wang, Y., Wang, Z., Xu, B., Du, Y., Lin, K., Xiao, Z., Yue, Z., Ju, J., Zhang, L., Yang, D., et~al.: Time-r1: Post-training large vision language model for temporal video grounding. arXiv preprint arXiv:2503.13377  (2025)

\bibitem{par_cvpr25}
Wang, Y., Ren, S., Lin, Z., Han, Y., Guo, H., Yang, Z., Zou, D., Feng, J., Liu, X.: Parallelized autoregressive visual generation. In: Proceedings of the IEEE/CVF Conference on Computer Vision and Pattern Recognition. pp. 12955--12965 (2025)

\bibitem{e2dvc_cvpr25}
Wu, K., Li, P., Fu, J., Li, Y., Wu, Y., Liu, Y., Wang, J., Zhou, S.: Event-equalized dense video captioning. In: Proceedings of the IEEE/CVF Conference on Computer Vision and Pattern Recognition. pp. 8417--8427 (2025)

\bibitem{moviebench_cvpr25}
Wu, W., Liu, M., Zhu, Z., Xia, X., Feng, H., Wang, W., Lin, K.Q., Shen, C., Shou, M.Z.: Moviebench: A hierarchical movie level dataset for long video generation. In: Proceedings of the Computer Vision and Pattern Recognition Conference. pp. 28984--28994 (2025)

\bibitem{moiveagent_arxiv25}
Wu, W., Zhu, Z., Shou, M.Z.: Automated movie generation via multi-agent cot planning. arXiv preprint arXiv:2503.07314  (2025)

\bibitem{qwen25omni}
Xu, J., Guo, Z., He, J., Hu, H., He, T., Bai, S., Chen, K., Wang, J., Fan, Y., Dang, K., et~al.: Qwen2. 5-omni technical report. arXiv preprint arXiv:2503.20215  (2025)

\bibitem{qwen3_omni_arxiv25}
Xu, J., Guo, Z., Hu, H., Chu, Y., Wang, X., He, J., Wang, Y., Shi, X., He, T., Zhu, X., et~al.: Qwen3-omni technical report. arXiv preprint arXiv:2509.17765  (2025)

\bibitem{streamingvlm_arxiv25}
Xu, R., Xiao, G., Chen, Y., He, L., Peng, K., Lu, Y., Han, S.: Streamingvlm: Real-time understanding for infinite video streams. arXiv preprint arXiv:2510.09608  (2025)

\bibitem{vid2seq_cvpr23}
Yang, A., Nagrani, A., Seo, P.H., Miech, A., Pont-Tuset, J., Laptev, I., Sivic, J., Schmid, C.: Vid2seq: Large-scale pretraining of a visual language model for dense video captioning. In: Proceedings of the IEEE/CVF conference on computer vision and pattern recognition. pp. 10714--10726 (2023)

\bibitem{mirage_latent_reasoning_cvpr26}
Yang, Z., Yu, X., Chen, D., Shen, M., Gan, C.: Machine mental imagery: Empower multimodal reasoning with latent visual tokens. In: Proceedings of the IEEE/CVF Conference on Computer Vision and Pattern Recognition. pp. 33510--33520 (2026)

\bibitem{timeexpert_iccv25}
Yang, Z., Yu, Y., Zhao, Y., Lu, S., Bai, S.: Timeexpert: An expert-guided video llm for video temporal grounding. In: Proceedings of the IEEE/CVF International Conference on Computer Vision. pp. 24286--24296 (2025)

\bibitem{dream_vl_video_dllm_arxiv25}
Ye, J., Gong, S., Gao, J., Fan, J., Wu, S., Bi, W., Bai, H., Shang, L., Kong, L.: Dream-vl \& dream-vla: Open vision-language and vision-language-action models with diffusion language model backbone. arXiv preprint arXiv:2512.22615  (2025)

\bibitem{llada_v}
You, Z., Nie, S., Zhang, X., Hu, J., Zhou, J., Lu, Z., Wen, J.R., Li, C.: Llada-v: Large language diffusion models with visual instruction tuning. arXiv preprint arXiv:2505.16933  (2025)

\bibitem{dllm_survey_arxiv25}
Yu, R., Li, Q., Wang, X.: Discrete diffusion in large language and multimodal models: A survey. arXiv preprint arXiv:2506.13759  (2025)

\bibitem{d2vlm_iccv25}
Zeng, W., Gao, D., Shou, M.Z., Ng, H.T.: Factorized learning for temporally grounded video-language models. In: Proceedings of the IEEE/CVF International Conference on Computer Vision. pp. 20683--20693 (2025)

\bibitem{timesuite_iclr25}
Zeng, X., Li, K., Wang, C., Li, X., Jiang, T., Yan, Z., Li, S., Shi, Y., Yue, Z., Wang, Y., et~al.: Timesuite: Improving mllms for long video understanding via grounded tuning. In: International Conference on Learning Representations (2025)

\bibitem{video_llama_emnlp23_demo}
Zhang, H., Li, X., Bing, L.: Video-llama: An instruction-tuned audio-visual language model for video understanding. In: Proceedings of the 2023 conference on empirical methods in natural language processing: system demonstrations. pp. 543--553 (2023)

\bibitem{timelens_cvpr26}
Zhang, J., Wang, T., Ge, Y., Ge, Y., Li, X., Shan, Y., Wang, L.: Timelens: Rethinking video temporal grounding with multimodal llms. arXiv preprint arXiv:2512.14698  (2025)

\bibitem{lpd_iclr26}
Zhang, Z., Huang, L.J., Wu, C., Yang, S., Peng, K., Lu, Y., Han, S.: Locality-aware parallel decoding for efficient autoregressive image generation. In: International Conference on Learning Representations (2026)

\bibitem{youcook2_aaai18}
Zhou, L., Xu, C., Corso, J.: Towards automatic learning of procedures from web instructional videos. In: Proceedings of the AAAI conference on artificial intelligence (2018)

\bibitem{masked_transformer_dvc_cvpr18}
Zhou, L., Zhou, Y., Corso, J.J., Socher, R., Xiong, C.: End-to-end dense video captioning with masked transformer. In: Proceedings of the IEEE conference on computer vision and pattern recognition. pp. 8739--8748 (2018)

\end{thebibliography}


\begin{thebibliography}{1}
\providecommand{\url}[1]{\texttt{#1}}
\providecommand{\urlprefix}{URL }
\providecommand{\doi}[1]{https://doi.org/#1}

\bibitem{qwen25vl}
Bai, S., Chen, K., Liu, X., Wang, J., Ge, W., Song, S., Dang, K., Wang, P., Wang, S., Tang, J., et~al.: Qwen2. 5-vl technical report. arXiv preprint arXiv:2502.13923  (2025)

\bibitem{chronusomni_arxiv25}
Chen, Y., Wu, Y., Guan, K., Ren, Y., Wang, Y., Song, R., Ru, L.: Chronusomni: Improving time awareness of omni large language models. arXiv preprint arXiv:2512.09841  (2025)

\bibitem{flashattention2}
Dao, T.: Flashattention-2: Faster attention with better parallelism and work partitioning. In: International Conference on Learning Representations. vol.~2024, pp. 35549--35562 (2024)

\bibitem{longvale}
Geng, T., Zhang, J., Wang, Q., Wang, T., Duan, J., Zheng, F.: Longvale: Vision-audio-language-event benchmark towards time-aware omni-modal perception of long videos. In: Proceedings of the Computer Vision and Pattern Recognition Conference. pp. 18959--18969 (2025)

\bibitem{etbench_nips24}
Liu, Y., Ma, Z., Qi, Z., Wu, Y., Shan, Y., Chen, C.W.: {E.T. Bench}: Towards open-ended event-level video-language understanding. In: The Thirty-eighth Conference on Neural Information Processing Systems Datasets and Benchmarks Track (2024)

\bibitem{video_salmonn2}
Tang, C., Li, Y., Yang, Y., Zhuang, J., Sun, G., Li, W., Ma, Z., Zhang, C.: video-salmonn 2: Captioning-enhanced audio-visual large language models. arXiv preprint arXiv:2506.15220  (2025)

\bibitem{qwen25omni}
Xu, J., Guo, Z., He, J., Hu, H., He, T., Bai, S., Chen, K., Wang, J., Fan, Y., Dang, K., et~al.: Qwen2. 5-omni technical report. arXiv preprint arXiv:2503.20215  (2025)

\bibitem{qwen3_omni_arxiv25}
Xu, J., Guo, Z., Hu, H., Chu, Y., Wang, X., He, J., Wang, Y., Shi, X., He, T., Zhu, X., et~al.: Qwen3-omni technical report. arXiv preprint arXiv:2509.17765  (2025)

\bibitem{d2vlm_iccv25}
Zeng, W., Gao, D., Shou, M.Z., Ng, H.T.: Factorized learning for temporally grounded video-language models. In: Proceedings of the IEEE/CVF International Conference on Computer Vision. pp. 20683--20693 (2025)

\end{thebibliography}
\end{document}